\documentclass[11pt]{article}

\usepackage[final]{acl}

\usepackage{times}
\usepackage{latexsym}

\usepackage[T1]{fontenc}

\usepackage[utf8]{inputenc}

\usepackage{microtype}

\usepackage{inconsolata}

\usepackage{graphicx}
\usepackage[most]{tcolorbox}
\usepackage{pifont}
\usepackage{booktabs}
\usepackage{multirow}
\usepackage{textcomp}
\usepackage{amssymb}
\usepackage[table]{xcolor}
\usepackage{hyperref}

\title{Beyond Literal Mapping: Benchmarking and Improving \\ Non-Literal Translation Evaluation}

\author{
Yanzhi Tian\textsuperscript{1}\thanks{Equal Contribution.}\thanks{Work was done when interned at Z.ai.} ~~~ Cunxiang Wang\textsuperscript{2,3}\footnotemark[1] ~~~ Zeming Liu\textsuperscript{4} ~~~ Heyan Huang\textsuperscript{1} \\ ~~~ {\bf Wenbo Yu\textsuperscript{2}} ~~~ {\bf Dawei Song\textsuperscript{1}} ~~~ {\bf Jie Tang\textsuperscript{3}} ~~~ {\bf Yuhang Guo\textsuperscript{1}}\thanks{Corresponding Author.} \\
\textsuperscript{1}School of Computer Science and Technology, Beijing Institute of Technology \\
\textsuperscript{2}Zhipu AI ~~~
\textsuperscript{3}The Knowledge Engineering Group (KEG), Tsinghua University \\
\textsuperscript{4}School of Computer Science and Engineering, Beihang University \\
\texttt{\normalsize \{tianyanzhi,guoyuhang\}@bit.edu.cn;} \texttt{\normalsize wangcunxiang303@gmail.com}; \texttt{\normalsize zmliu@buaa.edu.cn}
}

\begin{document}
\maketitle
\begin{abstract}
Large Language Models (LLMs) have significantly advanced Machine Translation (MT), applying them to linguistically complex domains-such as Social Network Services, literature etc. 
In these scenarios, translations often require handling non-literal expressions, leading to the inaccuracy of MT metrics.
To systematically investigate the reliability of MT metrics, we first curate a meta-evaluation dataset focused on non-literal translations, namely \textbf{MENT}.
MENT encompasses four non-literal translation domains and features source sentences paired with translations from diverse MT systems, with 7,530 human-annotated scores on translation quality.
Experimental results reveal the inaccuracies of traditional MT metrics and the limitations of LLM-as-a-Judge, particularly the knowledge cutoff and score inconsistency problem.
To mitigate these limitations, we propose RATE, a novel agentic translation evaluation framework, centered by a reflective Core Agent that dynamically invokes specialized sub-agents.
Experimental results indicate the efficacy of RATE, 
achieving an improvement of at least 3.2 points in combined system- and segment-level correlation with human judgments compared with current methods.
Further experiments demonstrate the robustness of RATE to general-domain MT evaluation.
Code and dataset are available at: \url{https://github.com/BITHLP/RATE}.
\end{abstract}

\section{Introduction}

Large Language Models (LLMs) have recently demonstrated remarkable capabilities across a wide range of NLP tasks \citep{2025-surveyllms}, and machine translation (MT) stands out as a particularly important application, not only because of its long-standing history in NLP, but also due to its practical significance in enabling global communication \citep{2024-MMT-llm, 2024-alma}.
LLMs have significantly expanded the application scope of MT, with the translation of online content—such as Social Network Services (SNS) on X, Facebook, and Rednote—emerging as a critical research direction. Effectively translating such content necessitates that models move beyond literal mapping to deeply understand internet slang, cross-cultural idioms, and literary contents \citep{2024-camt, 2025-seedx, 2025-redtrans, 2025-poetmt}.

\begin{figure}[t]
    \centering
    \includegraphics[width=0.98\linewidth]{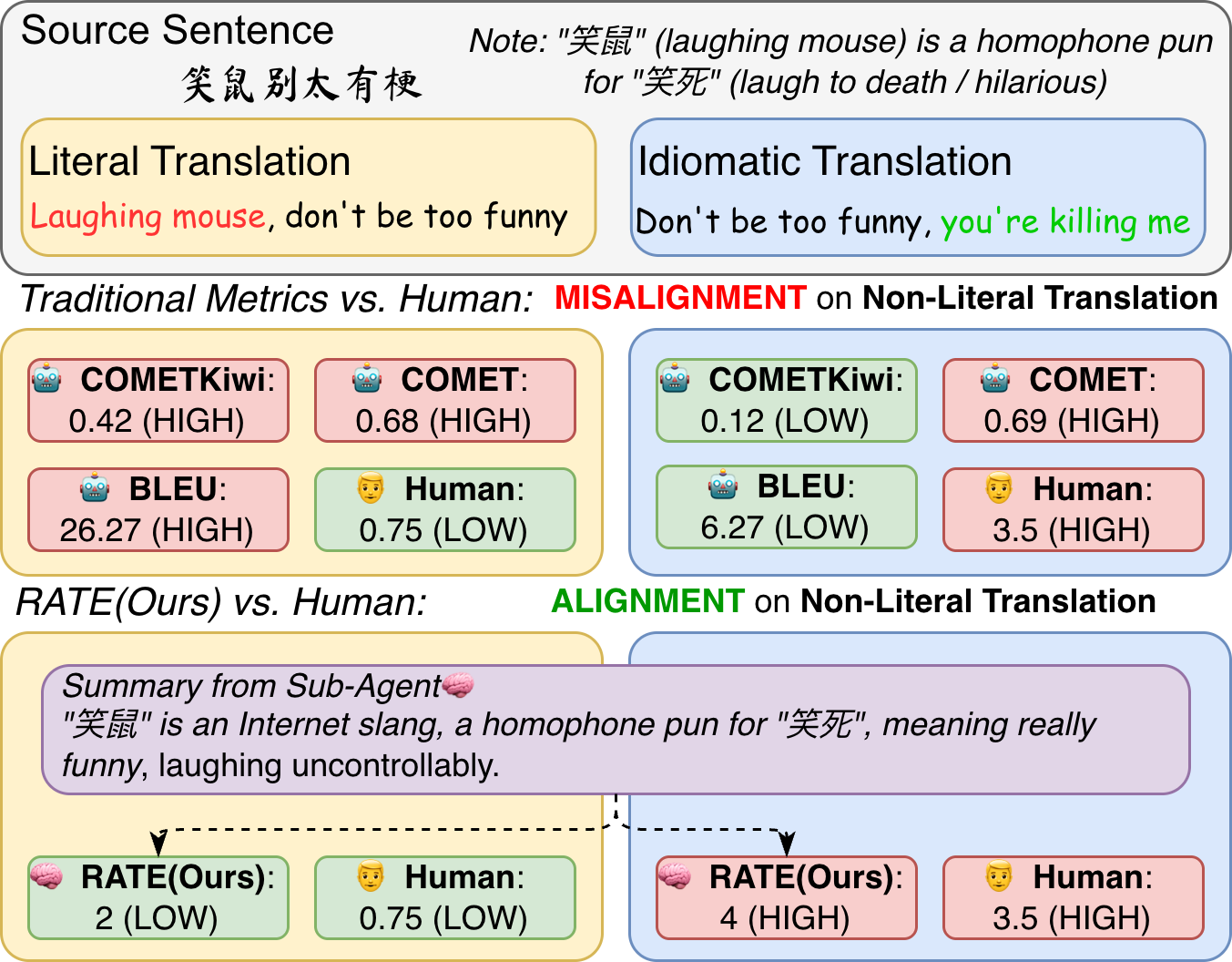}
    \caption{An illustration of evaluation misalignment in traditional metrics whereas the alignment with RATE. Traditional metrics often over-score literal but semantically incorrect translation (left) and penalize idiomatic translation (right). In contrast, our RATE leverages sub-agents to retrieve background knowledge and score calibration, achieving alignment with human judgment.}
    \label{fig:example}
\end{figure}

Given the complexities inherent in these non-literal translated scenarios, evaluating translation quality also becomes a challenge.
A precise assessment of translation quality requires relevant metrics that enable a deep understanding of both the source sentence and the translated sentence.
Through example from preliminary experiment in Figure \ref{fig:example}, it can be observed that evaluation using traditional MT metrics for translations in such domains fails to fully understand the complex sentence, resulting in the evaluation misalignment.
 
Systematically investigating the misalignment of MT metrics necessitates the construction of meta-evaluation dataset \citep{2019-takingMTevaluation, 2025-aces}.
Meta-evaluation is used to evaluate the correlation between MT automatic metrics and human judgment, which plays a crucial role in ensuring the accuracy and reliability of the metrics.
However, existing datasets used for evaluating these metrics typically contain source and translated texts which majority comes from conventional domains such as news and Wikipedia \citep{2022-demetr, 2023-wmtmetrics, 2025-aces}. These datasets lack exploration in the aforementioned non-literal domains raises concerns about the reliability of current MT metrics.

To assess the reliability of MT metrics, we firstly curate MENT, a \textbf{m}eta-\textbf{e}valuation dataset characterized by \textbf{n}on-literally \textbf{t}ranslation, that the source sentences are collected from SNS, cross-culture, poetry, and literature domains.
Our dataset comprises 7,530 human-annotated scores on translation quality, which allows for a systematic evaluation of current MT metrics \citep{2002-bleu, 2020-comet}, including LLM-as-a-Judge paradigms \citep{2023-gemba, 2024-eaprompt}.
Specifically, the collected source sentences are paired with several translations obtained from various MT systems, ranging from traditional NMT model to LLMs \citep{2022-nllb, 2025-seedx, 2025-qwen3, 2025-gemini25}.
Each translation is evaluated by at least two annotators, with quality control to ensure the reliability of the annotations.
Our evaluation results indicate that while LLM-as-a-Judge generally outperforms traditional MT metrics on the constructed dataset, but it still exhibits notable inaccuracies due to the inherent limitations of LLMs such as knowledge cutoffs, which affects the accurate assessment of recently emerged slang or evolving cultural expression.

To overcome the limitations of static LLM-as-a-Judge methods, particularly the knowledge cutoff and score inconsistency encountered when evaluating non-literal translations, we introduce a dynamic reflective procedure and propose a novel agentic framework RATE (\textbf{R}eflective \textbf{A}gentic \textbf{T}ranslation \textbf{E}valuation).
To achieve the reflective evaluation framework, RATE is powered by a Core Agent, which dynamically determines the selection of specialized sub-agents through its reasoning process.
It decides whether to invoke the Search Agent for external knowledge retrieval, delivering what kind of instruction and retrieved context to get a score from Evaluation Agent, or calibrating the evaluated score from Comparison Agent.
Experimental results on MENT dataset indicate the efficacy of RATE, 
which achieves an improvement of at least 3.2 points in combined system- and segment-level correlation with human judgments compared to current metrics including LLM-as-a-Judge for MT evaluation.
Further analysis experiments demonstrate the robustness of RATE, that its reliability is not limited to non-literal scenarios but also extends to general-domain MT evaluation.

Our contributions are summarized as follows:
\begin{itemize}
    \item To the best of our knowledge, we are the first to identify the challenges in accurately evaluating non-literal translation quality, and we construct a human-annotated meta-evaluation dataset MENT to systematically assess MT evaluation metrics.
    \item Our comprehensive meta-evaluation reveals the unreliability of MT metrics on non-literal content. Traditional metrics are fundamentally limited by the lack of deep semantic understanding, while LLM-as-a-Judge paradigms are hindered by the static knowledge cutoff and inherent score inconsistency.
    \item We propose a novel agentic translation evaluation framework RATE, which enables the dynamic invoking of specialized sub-agents by Core Agent. Experimental results demonstrate that RATE enhances reliability in non-literal translation evaluation while maintaining robustness across general domains.
\end{itemize}

\section{Related Work}

\begin{figure*}[t]
    \centering
    \includegraphics[width=0.98\linewidth]{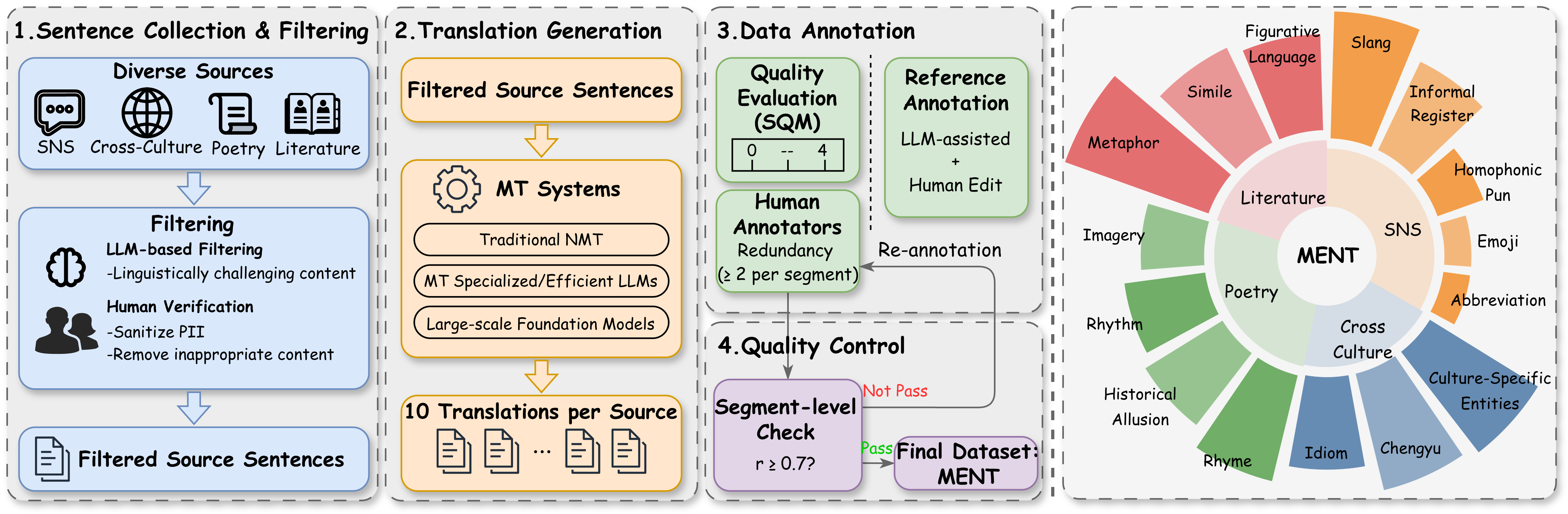}
    \caption{Overview of the data construction pipeline and final dataset visualization. \textbf{(Left)} The four-stage construction pipeline, proceeding from multi-domain sentences collection to strict quality control. \textbf{(Right)} Visualization of the MENT dataset, illustrating the distribution of source domains and linguistic phenomena.}
    \label{fig:data_construction}
\end{figure*}

\subsection{Machine Translation Metrics}
Accurately evaluating translation quality is crucial, as it not only enables reliable assessment and iteration of MT systems, but also provides trustworthy reward signals for preference optimization and reinforcement learning \citep{2025-mt-r1}.
Historically, rule-based metrics like BLEU \citep{2002-bleu} have been the standard evaluation metric across machine translation modalities \citep{2024-drda, 2024-dota, 2025-debackx, 2025-prim} due to efficiency and interpretability, yet they struggle to capture semantic lexical variability and diverse expressions \citep{2023-wmtmetrics}.
While model-based metrics \cite{2020-comet, 2023-metricx} improve semantic understanding through pre-trained encoders.
The recent emergence of LLM-as-a-judge methods \citep{2023-gemba, 2025-thinmqm} further leverages deep semantic reasoning capabilities of LLMs to achieve higher correlations with human judgments. Multi-agent frameworks \citep{2025-mad} attempt to extend this by coordinating multiple perspectives and hierarchical evaluation. 

\subsection{Machine Translation Meta-Evaluation}
The MT meta-evaluation, often referred to as ``evaluating MT metrics'', aims to validate the reliability of automatic metrics by measuring their alignment with human judgments. The majority of existing meta-evaluation datasets are constructed using literal or formal source texts \citep{2023-wmtmetrics, 2025-aces, 2022-demetr}, potentially overlooking complex linguistic phenomena.

Recent research explore the MT meta-evaluation in more challenging domains through small-scale human annotations.
\citet{2025-poetmt} manually evaluate 100 Chinese poetry translations, and \citet{2025-diting} annotate 300 Chinese web novel translations with multi-dimensional scores. 
Although these studies highlight the importance of extending MT meta-evaluation to more diverse and challenging domains, their annotation scale remains highly limited, with all datasets containing fewer than 1,000 annotated scores. 
Such small-scale datasets make it difficult to systematically validate the reliability and robustness of evaluation methods, which also underscores the inherent difficulty of constructing large-scale meta-evaluation datasets in these challenging domains.

\subsection{Agentic Evaluation}
Despite the success in capturing semantic nuances, conventional LLM judges often struggle with complex, multi-step reasoning tasks where reward signals are sparse or lack objective verifiability. To address these limitations, a new paradigm of ``Agent-as-a-Judge'' that employed agentic evaluation has emerged \citep{2026-agentasajudge}, where the evaluator functions as an autonomous agent capable of active reasoning and tool interaction.
\citet{2025-agenticrm} introduce Agentic Reward Modeling, which enhances the reliability of reward signals by integrating human preferences with verifiable correctness signals through an agentic workflow.
\citet{2025-evaluate-agent} demonstrate the efficacy of utilizing autonomous agents to verify hierarchical solution requirements, offering a scalable and cost-effective alternative to human evaluation.
Collectively, these advancements signify a pivotal shift toward more transparent, interpretable, and robust evaluation methodologies.

\section{Data Construction}
To systematically investigate the reliability of MT metrics, we construct \textbf{M}eta-\textbf{E}valuation dataset of \textbf{N}on-Literal \textbf{T}ranslation (MENT) following a four-stage pipeline, shown in Figure \ref{fig:data_construction}: (1) Sentence Collection and Filtering, (2) Translation Generation, (3) Data Annotation, and (4) Quality Control.

\subsection{Sentence Collection and Filtering}
Aligning with current research in MT meta-evaluation \citep{2022-demetr, 2025-aces}, we primarily collect source sentences from established MT benchmarks.
Specifically, we focus on four challenging domains: Social Network Services (SNS), Cross-Culture, Poetry, and Literature.
These domains are selected for their characterized with non-literal translation \citep{2025-redtrans, 2024-camt, 2025-poetmt, 2025-drt}. The following is a detailed description of the data sources for each domain.

\paragraph{Social Network Services (SNS) Domain:} The SNS domain is characterized by high-context linguistic phenomena, including Internet slang, abbreviations, homophonic puns, informal register etc. These features often defy literal translation constraints.
We collect English sentences from SNSBench \citep{2025-snsbench} and RedTrans \citep{2025-redtrans}, two datasets that contain user-generated content crawled from a social media, including user posted notes and comments, covering a wide range of topics.
To augment the dataset with more linguistically challenging Chinese social media texts, we crawl sentences from Chinese SNS platforms.

\paragraph{Cross-Culture Domain:} Texts in this domain feature culture-specific entities and idiomatic expressions that lack direct equivalents in the target language, necessitating cultural adaptation rather than literal translation.
For English, we source sentences from CAMT \citep{2024-camt}, an MT benchmark rich in culturally specific items.
For Chinese, we leverage ChID \citep{2019-chid}, a dataset designed for cloze tests focusing on Chinese idioms, Chengyu.

\paragraph{Poetry Domain:} Poetry presents unique challenges in preserving aesthetic features such as rhyme, rhythm, and imagery alongside semantic meaning, often requiring significant restructuring.
We crawl English poems from the Project Gutenberg repository.
Chinese poems are derived from PoetMT \citep{2025-poetmt}, encompassing Tang poetry and Song lyrics.

\paragraph{Literature Domain:} Literary translation demands the interpretation of figurative language, such as similes and metaphors, where surface-level translation often fails to convey the underlying intent.
Accordingly, English literary texts are collected from DRT \citep{2025-drt}, an MT dataset including English novels.
For Chinese, we leverage CMDAG \cite{2024-cmdag}, a multi-source dataset characterized by a high density of metaphorical expressions. We specifically extract samples from the prose subset of this corpus.

\paragraph{Data Filtering:} 
To ensure our collected data exhibits high translation difficulty, specifically targeting non-literal translation and adheres to strict privacy standards, we employ the data filtering strategy combining LLM-based scoring and human verification.
Initially, all collected source sentences undergo a preliminary filtering process using an LLM to retain linguistically challenging content (see Appendix \ref{sec:appendix_llm_filtering} for prompt details).
Subsequently, we conduct a manual inspection of the pre-filtered data, with particular attention to the newly crawled sentences. 
During this phase, we discard translation-irrelevant noise (e.g., topic hashtags in SNS content) and strictly sanitize the text to eliminate Personally Identifiable Information (PII), as well as offensive or inappropriate content.

\subsection{Translation Generation}

To construct a meta-evaluation dataset with diverse translation qualities and distinct error distributions, we employ several different MT systems to generate translations.
These systems span a wide spectrum of architectures and scales, ranging from traditional NMT to LLMs.
Specifically, we categorize the models into three groups based on their architecture and scale.
\textbf{Traditional NMT:} NLLB \cite{2022-nllb}.
\textbf{MT Specialized and Efficient LLMs:} Seed-X \cite{2025-seedx}, Hunyuan-MT \cite{2025-hunyuanmt}, Qwen3-8B \cite{2025-qwen3}, and Tower-Plus-9B \cite{2025-towerplus}. 
\textbf{Large-scale Foundation Models:} Qwen3-235B-A22B \cite{2025-qwen3}, GLM-4.5 \cite{2025-glm45}, DeepSeek-V3.1 \cite{2024-deepseekv3}, GPT-4o \cite{2024-gpt4o}, and Gemini-2.5Pro \cite{2025-gemini25}. 
Detailed specifications of each system are summarized in Table \ref{tab:models}.

\begin{table}[t]
\centering
\small
\resizebox{\columnwidth}{!}{%
\begin{tabular}{l|c|c|c}
\hline
\textbf{Model} & \textbf{Type} & \textbf{Params} & \textbf{Access} \\
\hline
\multicolumn{4}{l}{\textit{Traditional NMT}} \\
\hline
NLLB & Transformer & 3.3B & Open \\
\hline
\multicolumn{4}{l}{\textit{MT Specialized \& Efficient LLMs}} \\
\hline
Seed-X & MT Specialized & 7B & Open \\
Hunyuan-MT & MT Specialized & 7B & Open \\
Qwen3-8B & General LLM & 8B & Open \\
Tower-Plus-9B & MT Specialized & 9B & Open \\
\hline
\multicolumn{4}{l}{\textit{Large-scale Foundation Models}} \\
\hline
Qwen3-235B-A22B & General LLM & 235B & Open \\
GLM-4.5 & General LLM & 355B & Open \\
DeepSeek-V3.1 & General LLM & 671B & Open \\
GPT-4o & General LLM & - & Proprietary \\
Gemini-2.5Pro & General LLM & - & Proprietary \\
\hline
\end{tabular}%
}
\caption{Overview of MT systems used for translation generation, categorized by architecture and scale.}
\label{tab:models}
\end{table}

\subsection{Data Annotation}
All recruited annotators hold degrees in translation and possess extensive prior experience in translation evaluation tasks. 
\paragraph{Translation Quality.} We adopt the Scalar Quality Metrics (SQM) framework, a human evaluation protocol used in WMT tasks \citep{2021-wmtmqm}. 
The choice of the SQM protocol is primarily driven by the distinct characteristics of non-literal translation in MENT.
Unlike literal mapping, failures in non-literal contexts often stem from a holistic misunderstanding of the source text's slang or cultural nuances, rather than isolated lexical errors.
Consequently, identifying specific error spans under an annotation protocol like Multidimensional Quality Metrics (MQM) \citep{2013-mqm} becomes highly subjective, as the error often permeates the entire sentence structure.
Following previous practices on human evaluation \citep{2024-maple, 2025-seedx, 2025-hunyuanmt}, we employ a 5-point scale based on the specific criteria illustrated in Appendix \ref{sec:appendix_annotation}.

To ensure the reliability of the annotations, we introduce redundancy into the annotation process. Specifically, each segment (10 translations corresponding to a source sentence) is evaluated by at least two independent annotators, and the final quality score for each translation is derived by averaging these individual ratings.

\paragraph{Translation Reference.} To assess the reference-based metrics, we further manually annotate the translation reference. 
We utilize an LLM-assisted approach where each sentence is translated 5 times by Gemini-3.0Pro, one of the most advanced LLMs currently available, and annotators are asked to select and post edit the best translation.
During annotation, 29.0\% of references are post edited.
For the sentences collected from PoetMT \cite{2025-poetmt}, we adopt the references in the dataset, since these references are annotated by professional translator. Specifically, we perform manual sentence-level alignment and review of the references from original dataset.
For detailed annotation guideline, please refer to Appendix \ref{sec:appendix_annotation}.

\begin{figure}[t]
    \centering
    \includegraphics[width=0.98\linewidth]{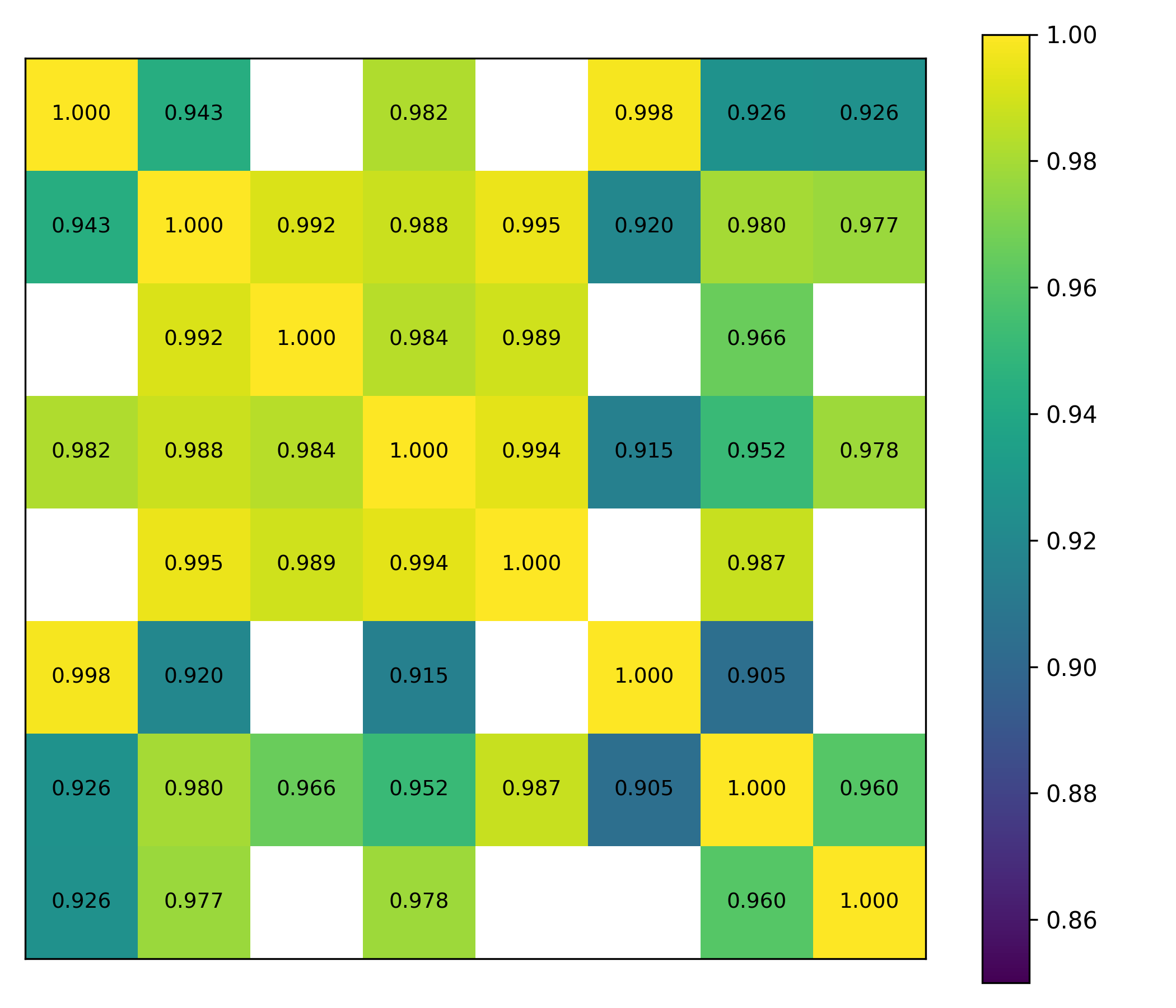}
    \caption{Heatmap of Pearson correlations for system-level inter-annotator agreement. Each cell $(i, j)$ displays the correlation coefficient calculated on the aggregated scores of shared translations. Blank cells denote pairs with no overlapping annotation tasks due to the workload distribution. 
    }
    \label{fig:annotator_correlation}
\end{figure}

\subsection{Quality Control}

\paragraph{Segment-level Agreement.}
To ensure quality control, we enforce a strict agreement threshold at the segment level.
Specifically, for each segment, the scores assigned by any pair of annotators across the translations must exhibit a Pearson correlation coefficient $r$ of at least 0.7.
If this threshold is not met, the segment is flagged as unreliable and sent for re-annotation.

During the quality control, 22.2\% of segments fail to meet the consistency criteria and subsequently undergo re-annotation to resolve discrepancies.
Among these, 4.7\% still remain inconsistent and are discarded as the annotators cannot  reach a consensus, thereby ensuring the high reliability of the final dataset.

\begin{table}[t]
\centering
\small
\begin{tabular}{c|c|c|c}
\toprule
 & \textbf{ZH-EN} & \textbf{EN-ZH} & \textbf{Total} \\
\midrule
SNS & 1,100 & 710 & 1,810 \\
Cross-Culture & 950 & 920 & 1,870 \\
Poetry & 1,060 & 840 & 1,900 \\
Literature & 870 & 1,080 & 1,950 \\
\midrule
Total & 3,980 & 3,550 & 7,530 \\
\bottomrule
\end{tabular}
\caption{Statistic of MENT, the numbers represent the count of translations (Segments $\times$ 10 MT Systems) collected for each domain and direction.}
\label{tab:data_statistic}
\end{table}

\subsection{Data Statistics and Quality}
Table \ref{tab:data_statistic} provides statistics of MENT. Samples of our annotated dataset are shown in Appendix \ref{sec:appendix_data_sample}.

To evaluate the quality of the human-assigned translation scores, we measure the \textbf{Inter-Annotator Agreement (IAA)} using Pearson correlation coefficient between annotators based on their shared workload.
As visualized in Figure \ref{fig:annotator_correlation}, the value at position $(i, j)$ in the heatmap represents the correlation coefficient between annotator $i$ and annotator $j$, calculated over the set of segments that they both annotate. 
Blank cells in the heatmap indicate pairs of annotators who do not share any overlapping assignments.
The high correlation values ($>0.9$) indicate high reliability of the annotated translation quality scores.
Detailed calculation method of IAA is introduced in Appendix \ref{sec:appendix_IAA}.
 
The quality of human-annotated references is further verified through a \textbf{secondary review process}, achieving a pass rate of 96.1\%, thereby guaranteeing the quality of annotated references.

\section{RATE: Reflective Agentic Translation Evaluation}
\label{sec:rate}

To mitigate the limitations in current metrics for MT evaluation, especially to the knowledge cutoff and score inconsistency of LLM-as-a-Judge, 
we propose an agentic evaluation framework \textbf{R}eflective \textbf{A}gentic \textbf{T}ranslation \textbf{E}valuation (RATE).

\begin{figure}[t]
    \centering
    \includegraphics[width=0.98\linewidth]{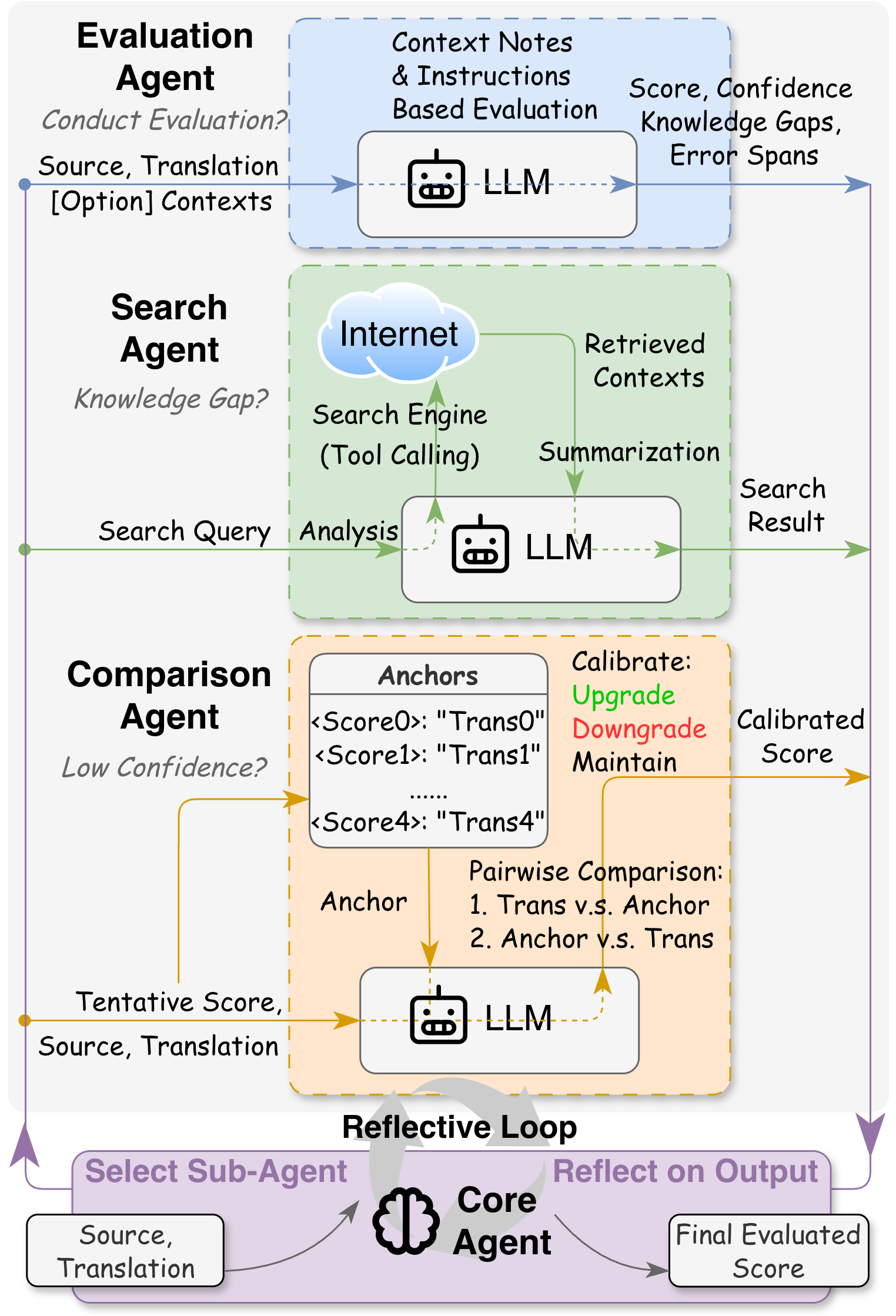}
    \caption{Overview of the RATE framework. The Core Agent acts as the central controller, dynamically selecting specialized sub-agents (Evaluation, Search, and Comparison) based on current state and outputs of sub-agents to iteratively refine the translation evaluation.}
    \label{fig:rate_framework}
\end{figure}

In contrast to existing LLM-based paradigms for MT evaluation \citep{2023-gemba, 2025-mad, 2025-himate}, 
RATE is not constrained by a static workflow with fixed, sequential evaluation procedures, or limited external tool usage (e.g., search engine). It is architected around a centralized \textbf{Core Agent}, which orchestrates three functional sub-agents: the \textbf{Evaluation Agent} for pointwise assessment, the \textbf{Search Agent} for online knowledge retrieval, and the \textbf{Comparison Agent} for calibration by pairwise evaluation, as shown in Figure \ref{fig:rate_framework}.

\subsection{Architectures}

\paragraph{Core Agent.} The Core Agent serves as the central controller of the framework. Unlike a simple router, it operates on a reflective loop based on OODA (Observe, Orient, Decide, and Act). 
Tracking the current understanding of the source sentence, the accumulated knowledge, and the confidence level of current judgments, it automatically determines whether to request a score (via Evaluation), retrieve more information (via Search), or verify a judgment (via Comparison).
This process operates iteratively, and the Core Agent orchestrates multi-turn interactions with sub-agents to continuously refine its assessment. The loop terminates only when the agent determines that sufficient evidence has been gathered to output a final evaluation, or when a pre-defined maximum round limit is reached.

\paragraph{Evaluation Agent.} This agent functions as the primary pointwise assessor, tasked with analyzing the translation quality. 
Crucially, it is designed to accept the optional context notes and instructions, with explanations of necessary knowledge injected from Core Agent, to ensure the evaluation is not based on hallucinations.
Furthermore, beyond producing a scalar score with confidence and scoring rationale, this agent is asked to return the error spans and suspected knowledge gaps, thereby triggering the Core Agent to activate other agent for clarification.

\begin{table*}[t]
\centering
\small
\begin{tabular}{l c cccccc cccccc}
\toprule
 & & \multicolumn{6}{c}{\textbf{ZH-EN}} & \multicolumn{6}{c}{\textbf{EN-ZH}} \\
\cmidrule(lr){3-8} \cmidrule(lr){9-14}

\multirow{2}{*}{\textbf{Metric}} & \multirow{2}{*}{\textbf{Meta}} & \multicolumn{3}{c}{System-Level} & \multicolumn{3}{c}{Segment-Level} & \multicolumn{3}{c}{System-Level} & \multicolumn{3}{c}{Segment-Level} \\
\cmidrule(lr){3-5} \cmidrule(lr){6-8} \cmidrule(lr){9-11} \cmidrule(lr){12-14}

 & & Acc. & $r$ & $\rho$ & Acc-t. & $r$ & $\rho$ & Acc. & $r$ & $\rho$ & Acc-t. & $r$ & $\rho$ \\
\midrule

\rowcolor{gray!15}
\multicolumn{14}{l}{\textit{Reference-based Metrics}} \\
BLEU & 64.9 & 82.2 & 91.1 & 84.2 & 52.0 & 30.1 & 30.7 & 91.1 & 92.6 & 91.5 & \cellcolor{gray!30}{\textbf{56.6}} & 37.0 & 39.5 \\
BLEURT & 56.5 & \cellcolor{gray!30}{\textbf{88.9}} & \cellcolor{orange!25}{\textbf{98.4}} & \cellcolor{gray!30}{\textbf{90.3}} & 55.4 & 45.4 & \cellcolor{gray!30}{\textbf{40.5}} & 57.8 & 82.5 & 17.6 & 50.5 & 28.5 & 21.8 \\
COMET & \cellcolor{gray!30}{\textbf{69.0}} & \cellcolor{orange!25}{\textbf{88.9}} & \cellcolor{gray!30}{\textbf{98.2}} & \cellcolor{orange!25}{\textbf{91.5}} & \cellcolor{gray!30}{\textbf{56.5}} & 50.3 & 43.3 & 82.2 & 92.0 & 79.4 & 56.5 & 50.3 & 38.8 \\
XCOMET & 54.5 & 80.0 & 96.7 & 73.3 & 48.9 & 29.3 & 21.2 & 68.9 & 86.6 & 40.6 & 52.0 & 31.1 & 25.8 \\
MetricX-23 & 52.2 & 73.3 & 94.7 & 62.4 & 47.5 & 31.5 & 17.3 & 64.4 & 83.6 & 27.3 & 53.2 & 41.2 & 29.4 \\
MetricX-24  & 56.2 & 73.3 & 94.6 & 55.2 & 53.8 & 47.1 & 35.9 & 66.7 & 83.5 & 29.7 & 54.5 & 47.1 & 33.0 \\
\midrule

\rowcolor{gray!15}
\multicolumn{14}{l}{\textit{Reference-free (QE) Metrics}} \\
COMETKiwi & 42.8 & 51.1 & 90.6 & 5.5 & 52.3 & 38.6 & 31.6 & 48.9 & 79.5 & -1.8 & 52.3 & 37.8 & 26.9 \\
MetricX23-QE & 31.9 & 46.7 & 86.4 & -5.5 & 41.9 & 15.2 & 0.4 & 35.6 & 79.4 & -28.5 & 50.2 & 39.9 & 20.8 \\
MetricX24-QE & 41.7 & 48.9 & 88.4 & -1.8 & 47.9 & 31.7 & 18.6 & 55.6 & 79.3 & 12.7 & 51.8 & 41.5 & 25.4 \\
\midrule

\rowcolor{gray!15}
\multicolumn{14}{l}{\textit{LLM-as-a-Judge}} \\
GEMBA-MQM & 68.7 & 82.2 & 97.0 & 80.6 & 46.2 & \cellcolor{gray!30}{\textbf{56.0}} & 26.8 & \cellcolor{orange!25}{\textbf{91.1}} & \cellcolor{orange!25}{\textbf{97.2}} & \cellcolor{orange!25}{\textbf{93.9}} & 52.1 & \cellcolor{gray!30}{\textbf{58.2}} & 42.6 \\
GEMBA-DA & \cellcolor{orange!25}{\textbf{77.2}\cellcolor{orange!25}} & 86.7 & 96.5 & 87.9 & \cellcolor{orange!25}{\textbf{59.1}} & \cellcolor{orange!25}{\textbf{70.1}} & \cellcolor{orange!25}{\textbf{60.2}} & \cellcolor{purple!30}{\textbf{91.1}} & 94.5 & \cellcolor{purple!30}{\textbf{95.2}} & \cellcolor{orange!25}{\textbf{56.7}} & \cellcolor{purple!30}{\textbf{68.4}} & \cellcolor{purple!30}{\textbf{60.5}} \\  
EAPrompt & 47.4 & 82.2 & 93.6 & 74.5 & 36.1 & -6.4 & -10.9 & 75.6 & 92.4 & 67.3 & 43.3 & 12.2 & 9.1 \\
ThinMQM & 65.1 & 80.0 & 95.1 & 80.6 & 49.5 & 34.8 & 29.8 & \cellcolor{gray!30}{\textbf{91.1}} & 91.5 & 91.5 & 50.4 & 43.8 & 43.2  \\
M-MAD & 65.8 & 82.2 & 96.3  & 83.3 & 43.9 & 39.6 & 23.0 & 88.9 & \cellcolor{gray!30}{\textbf{95.2}} & 90.3 & 51.5 & 50.1 & \cellcolor{gray!30}{\textbf{45.2}} \\
\textbf{RATE (Ours)} & \cellcolor{purple!30}{\textbf{80.4}} & \cellcolor{purple!30}{\textbf{97.8}} & \cellcolor{purple!30}{\textbf{99.3}} & \cellcolor{purple!30}{\textbf{99.7}} & \cellcolor{purple!30}{\textbf{61.9}} & \cellcolor{purple!30}{\textbf{74.5}} & \cellcolor{purple!30}{\textbf{66.4}} & 88.9 & \cellcolor{purple!30}{\textbf{97.7}} & \cellcolor{gray!30}{\textbf{92.7}} & \cellcolor{purple!30}{\textbf{59.5}} & \cellcolor{orange!25}{\textbf{65.3}} & \cellcolor{orange!25}{\textbf{60.1}} \\
\bottomrule
\end{tabular}
\caption{System-level and segment-level correlations on the MENT dataset. We report \textbf{Accuracy (Acc., Acc-t.)}, \textbf{Pearson($r$)}, and \textbf{Spearman ($\rho$)} correlation coefficients, scaled by a factor of 100. Meta represents the average score of all accuracies and correlation coefficients. The \colorbox{purple!30}{best}, \colorbox{orange!25}{second-best}, and \colorbox{gray!30}{third-best} results are marked with purple, orange, and gray backgrounds respectively.}
\label{tab:main_results}
\end{table*}

\paragraph{Search Agent.} This agent is used to mitigate the ``knowledge cutoff'' limitation inherent in LLMs.
It is invoked on-demand by the Core Agent when the source text contains ambiguous entities, emerging internet slang, or deep cultural idioms that cannot be resolved via internal parametric knowledge of LLMs.
The Search Agent first analyzes request from Core Agent, and then transfer it into search query to call the search engine. After obtaining several responses from search engine, it summarizes the related responses and return it to Core Agent.

\paragraph{Comparison Agent.} This agent serves as a critical calibration module. It is specifically designed to mitigate the inherent subjectivity and potential inaccuracies associated with the pointwise evaluation paradigm employed by the Evaluation Agent, with utilization previously evaluated translations as anchors. When the Core Agent is uncertain about a new translation's quality, it instructs this agent to perform a pairwise preference ranking against these anchors.
This mechanism converts subjective absolute judgments into robust relative rankings, mitigating score inconsistency of the evaluation.

The detail implementations of RATE are introduced in Appendix \ref{sec:appendix_rate_impl}. We provide evaluation trajectories of RATE, shown in Appendix \ref{sec:appendix_rate_sample}.

\section{Experiments}
\subsection{Experimental Setup}
\paragraph{Meta-Evaluation.} Following WMT23 Metrics Shared Task \cite{2023-wmtmetrics}, we adopt the composite meta score to assess metric performance across both system level and segment level.
Our evaluation includes Accuracy (Acc \citep{2021-acc}, Acc-t \citep{2023-acc-t}), Pearson Correlation ($r$), and Spearman correlation ($\rho$). The final meta score is calculated as the average of individual statistics \citep{2025-mad}.
All results are computed based on MTME, the standard metric evaluation tool recommended by WMT.

\paragraph{Evaluated Metrics.}
We evaluate a comprehensive set of metrics categorized into three paradigms:
Reference-based metrics, Reference-free (Quality Estimation, QE) metrics, and LLM-as-a-Judge adopting reference-free paradigm.
Detailed experimental setup is introduced in Appendix \ref{sec:appendix_exp_setup}.

\subsection{Main Results}
The experimental results are shown in Table \ref{tab:main_results}.

\paragraph{Reference-based Metrics.} The gold-standard reference essentially introduces the necessary semantic correspondences for non-literal expressions, thereby allowing the metrics to maintain a certain level of reliability. 
As observed in Table \ref{tab:main_results}, methods within this category perform competitively, with COMET achieving the highest meta score (69.0) among reference-based metrics.

\paragraph{Breakdown of Reference-free (QE) Metrics.} 
Most QE models rely on pre-trained multilingual encoders, which are primarily trained on literal or formal corpora, and cannot verify the quality of non-literal translation. 
As clearly reflected in Table \ref{tab:main_results}, this leads to a significant performance breakdown: QE metrics yield the lowest meta scores overall (ranging from 31.9 to 42.8).

\paragraph{The Power of Semantic Interpretation in LLM-as-a-Judge.} LLM-as-a-Judge paradigms demonstrate remarkable human alignment even without access to gold-standard references. For instance, GEMBA-DA achieves a strong meta score of 77.2 in Table \ref{tab:main_results}. 
This efficacy stems from the inherent capability of LLMs to perform deep semantic interpretation.
RATE further enhances the performance by dynamically invoking the sub-agents, and achieves the highest meta score of 80.4 among all evaluation methods.

The additional experimental results on system-level Soft Pairwise Accuracy (SPA) are shown in Appendix \ref{sec:appendix_spa}, further demonstrating the effectiveness of RATE in providing reliable system-level rankings for non-literal translations.

\subsection{Domain-Specific Performance Analysis}
We further investigate the performance of different evaluation paradigms with specific domain.
The experimental results are illustrated in Figure \ref{fig:domains_meta_score}, and more detailed results are shown in Appendix \ref{sec:appendix_domain_eval_result}.

\label{sec:domain_specific_analysis}
\begin{figure}[t]
    \centering
    \includegraphics[width=0.98\linewidth]{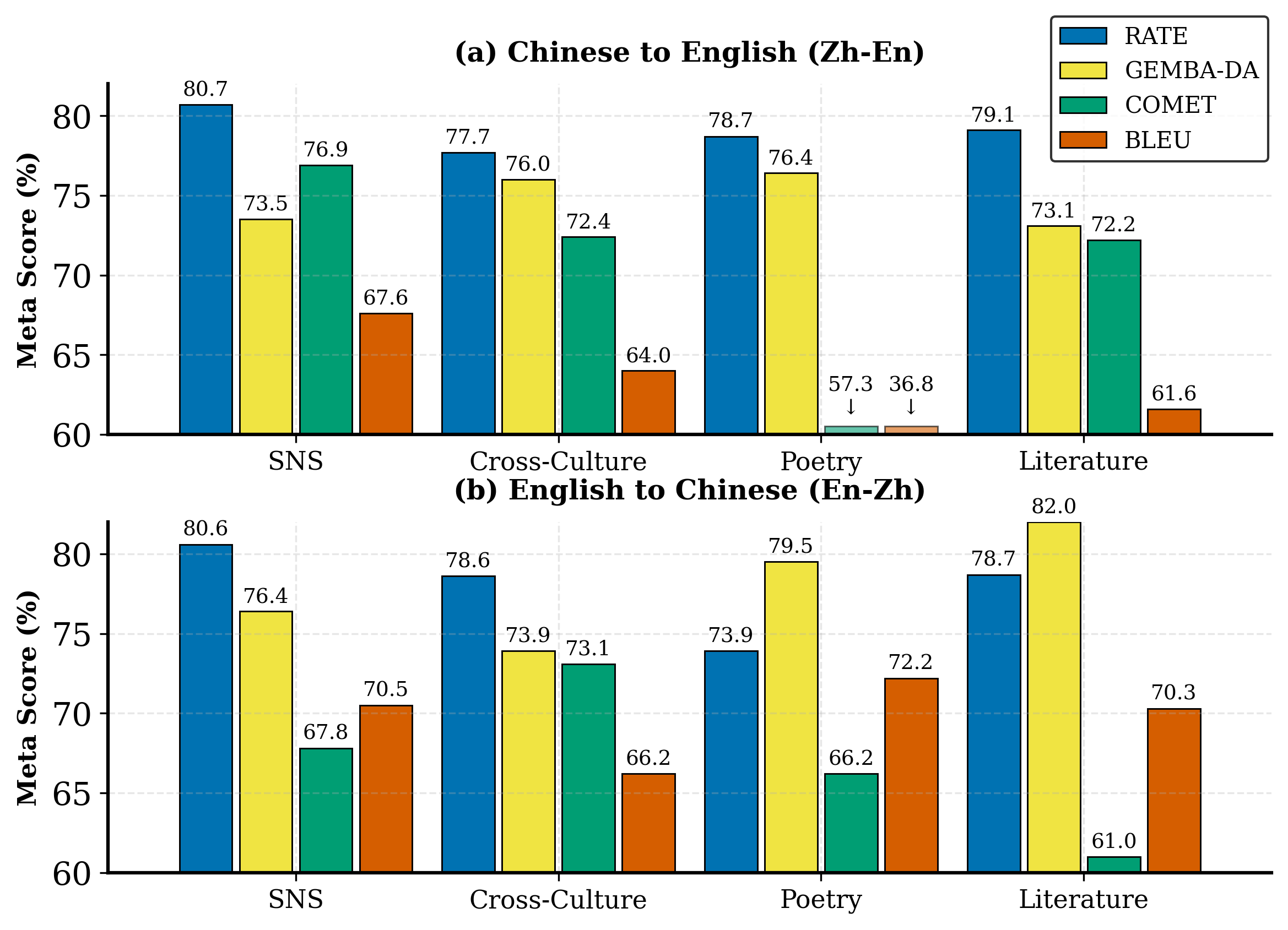}
    \caption{Illustration of metrics performance with specific domain, detailed numerical results in Appendix \ref{sec:appendix_domain_eval_result}.}
    \label{fig:domains_meta_score}
\end{figure}

\paragraph{SNS and Cross-Culture Domains:} The primary challenge of evaluating translations in these two domains lies in the utilization of  emerging slang, and idiomatic expressions. 
These linguistic phenomena are frequently absent from the static pre-training corpora of LLMs, leading to ``knowledge cutoff'' that hinders the performance of LLM-as-a-Judge paradigm.
Furthermore, the subtle semantic nuances distinguishing various translation candidates in these contexts can lead to significant distinction in overall quality, and LLMs often fail to assess with consistent scoring.
With dynamically invoking sub-agents, RATE enhances the understanding by external knowledge, or calibrates the score by pairwise comparison, thereby mitigates the limitations of current LLM-as-a-Judge methods.

\paragraph{Poetry and Literature Domains:} These domains focus on figurative complexities, such as metaphor and imagery, rather than currently evolving slang. In the En-Zh direction, current LLM-as-a-Judge methods already demonstrate superior performance, significantly outperforming traditional metrics by leveraging their deep internalized semantic reasoning.
However, evaluation in the Zh-En direction remains uniquely challenging due to the prevalence of historical allusions in Chinese poetry and literature. In such cases, Search Agent of RATE serves as a vital knowledge bridge, injecting necessary background context to decipher these allusions, ensuring that the evaluation is grounded in cultural accuracy rather than just linguistic fluency, further enhancing reliability where static LLMs reach their limits.

\section{Analysis}

\subsection{Temporal Dynamics and Behavioral Analysis of Sub-Agents Invoking}

We investigate the temporal and behavioral patterns of sub-agent invocations, with the distribution visualized in Figure \ref{fig:agent_distribution}. Detailed agentic evaluation trajectories are provided in Appendix \ref{sec:appendix_rate_sample}.

\begin{figure}[t]
    \centering
    \includegraphics[width=0.98\linewidth]{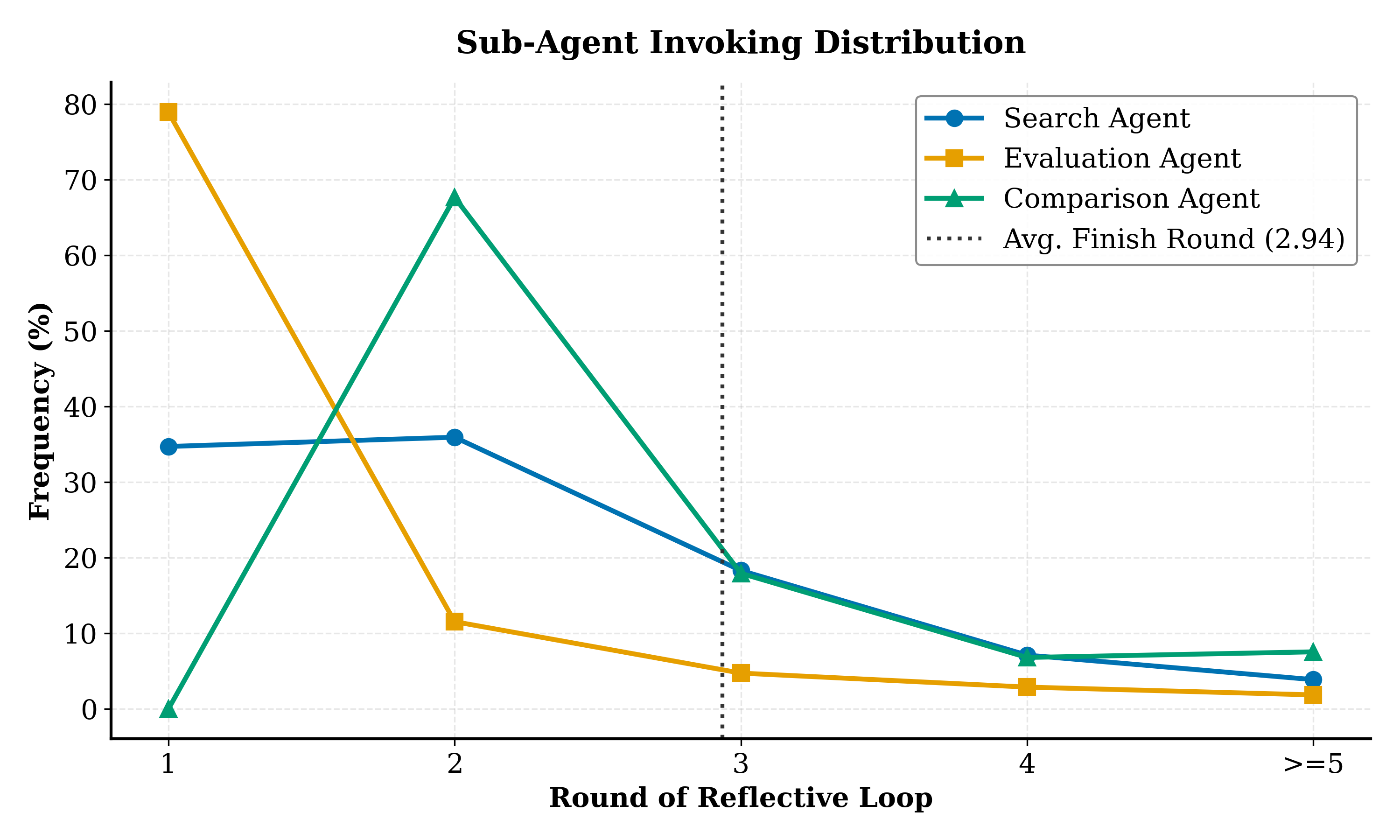}
    \caption{Temporal distribution of sub-agent invoking, illustrating the transition from initial evaluation to background knowledge-enhanced calibration.}
    \label{fig:agent_distribution}
\end{figure}

The invoking of the Search Agent is primarily concentrated in the early stages, where the Core Agent directly requests external knowledge to resolve non-literal translation content. However, it is also observed in later rounds. This indicates that feedback from the Evaluation Agent may reveal latent comprehension gaps that are not initially apparent, and the Core Agent asks the Search Agent for deeper clarification.

The Evaluation Agent is also mainly selected in the initial stages.
For less complex texts, the Core Agent firstly attempts invoking the Evaluation Agent, to determine if a high-confidence assessment can be achieved immediately.
For more complex texts, the Core Agent can leverage the accumulated context, allowing the Evaluation Agent to proceed with a knowledge-enhanced assessment. Detailed implementation of accumulation mechanism is provided in Appendix \ref{sec:appendix_rate_impl}.

Consistent with our design expectations, the Comparison Agent is not invoked in the first round, since the agent is triggered only when the Core Agent lacks confidence in the preliminary score provided by the Evaluation Agent, serving as a necessary mechanism to calibrate the score.

\subsection{Ablation Study}

To quantify the contributions of sub-agents in the reflective framework, we conduct an ablation study by restricting the Core Agent’s calling protocol (removing sub-agent calling in prompt).

\begin{table}[ht]
\centering
\small
\resizebox{\columnwidth}{!}{
    \begin{tabular}{l ccc}
    \toprule
    \multirow{2}{*}{\textbf{Setting}} & \multicolumn{3}{c}{\textbf{Meta}}  \\
    \cmidrule(lr){2-4}
    & ZH-EN & EN-ZH & Avg. \\
    \midrule
    RATE & 83.3 & 77.4 & 80.4 \\
    \quad w/o Search Agent & 78.2 \color{red}{(-5.1)} & 76.9 \color{red}{(-0.5)} & 77.6 \color{red}{(-2.8)} \\
    \quad w/o Comparison Agent & 81.7 \color{red}{(-1.6)} & 76.3 \color{red}{(-1.1)} & 79.0 \color{red}{(-1.4)} \\
    \quad w/o Both & 77.6 \color{red}{(-5.7)} & 76.0 \color{red}{(-1.4)} & 76.8 \color{red}{(-3.6)} \\
    \bottomrule
    \end{tabular}
}
\caption{Experimental results of ablation study. Specifically, we evaluate three variants: (1) w/o Search Agent; (2) w/o Comparison Agent; and (3) w/o Both, which simplifies the framework into only a Evaluation Agent cooperating with Core Agent.}
\label{tab:ablation}
\end{table}

As shown in Table \ref{tab:ablation}, the removal of Search Agent results in the most significant performance degradation, confirming for non-literal translation data, the static internalized parameters of LLM are insufficient. 
Disabling Comparison Agent also leads to decreasing of performance, indicating the efficacy of pairwise calibration.
The experimental results demonstrate the necessity of dynamic sub-agents utilization in RATE framework.

\subsection{Evaluation of RATE on General Dataset}
We conduct experiments on WMT23 En-De Metrics Shared Task, which represents a standard general domain meta-evaluation dataset, with new language direction, and significantly differs from our primary focus on non-literal translation scenarios.
Crucially, we \textbf{maintain the same system prompts and agentic protocols without any modifications}.

\begin{table}[ht]
\centering
\small
\resizebox{\columnwidth}{!}{%
    \begin{tabular}{l cc cc}
    \toprule
    \multirow{2}{*}{\textbf{Metric}} & \multicolumn{2}{c}{\textbf{System-Level}} & \multicolumn{2}{c}{\textbf{Segment-Level}} \\
    \cmidrule(lr){2-3} \cmidrule(lr){4-5}
    & Acc. & $r$ & Acc-t. & $r$ \\
    \midrule
    \rowcolor{gray!15}
    \multicolumn{5}{l}{\textit{Reference-based Metrics}} \\
    BLEU & 89.4 & 91.7 & 52.0 & 19.2  \\
    COMET & 97.0 & \textbf{99.0} & \underline{57.4} & 43.2 \\
    MetricX-23 & 90.9 & 97.7 & 60.3 & \underline{58.5} \\
    \midrule
    \rowcolor{gray!15}
    \multicolumn{5}{l}{\textit{Reference-free (QE) Metrics}} \\
    COMETKiwi &	\underline{98.5} & 94.6 & 56.9 & 47.5 \\
    MetricX23-QE & 92.4 & 96.7 & \textbf{60.3} & \textbf{62.6} \\
    \midrule
    \rowcolor{gray!15}
    \multicolumn{5}{l}{\textit{LLM-as-a-Judge}} \\
    EAPrompt$^*$ & 93.9 & 96.2 & 47.1 & 52.0 \\
    GEMBA-MQM$^*$ & 97.0 & 97.3 & 47.4 & 42.9 \\
    M-MAD$^*$ & 97.0 & 97.9 & 55.5 & 55.2 \\
    \textbf{RATE (Ours)} & \textbf{98.5} & \underline{99.0} & 52.3 & 38.9  \\
    \bottomrule
    \end{tabular}
}
\caption{Performance on the WMT23 En-De Metrics Shared Task \citep{2023-wmtmetrics}. $^*$ indicates the results are derived from \citet{2025-mad}.}
\label{tab:wmt23_ende_results}
\end{table}

Experimental results in Table \ref{tab:wmt23_ende_results} show that RATE achieves a comparable performance to metrics on system-level translation evaluation, indicating robustness of the proposed RATE.

\section{Conclusion}
In this paper, we identify the challenges in accurately evaluating non-literal translation. 
To systematically investigate the reliability of MT metrics, we construct a human-annotated meta-evaluation dataset focusing on non-literal translation, MENT.
Our comprehensive evaluation reveals the inaccuracy of MT metrics.
To mitigate the limitations of knowledge cutoff and score inconsistency of LLM-as-a-Judge methods, we propose a novel agentic translation evaluation framework RATE.
Experimental results demonstrate that RATE enhances reliability in non-literal translation evaluation, while further analysis experiments indicate its robustness across general domains.

\section*{Limitations}
While the RATE framework demonstrates improvement in translation evaluation, certain limitations remain regarding the sub-agent invoking within the reflective loop.

In the current architecture, specialized sub-agents and external search engines are invoked via a tool-calling mechanism. However, these external services are susceptible to execution failures, such as LLM API timeouts or search engine connectivity errors.
Presently, the Core Agent lacks the analytical capability to interpret diagnostic error messages or implement adaptive retry strategies.
When an execution failure of external services occurs, system defaults to re-initiating the evaluation process rather than performing a localized recovery or retry, which could potentially hinders the overall efficiency of the framework, particularly in scenarios where external API stability fluctuates.

\section*{Ethics Statement}
We take ethical considerations seriously and ensure that the data used in this study is conducted in a responsible and ethical manner.
The primary data sources for this research include both newly curated web-crawled content and existing open-source datasets.

For the crawled data, we perform rigorous manual inspection and filtering to ensure that the dataset contains no Personally Identifiable Information (PII) or sensitive user data, and we exclude any content that could be deemed offensive, harmful, or culturally insensitive.
Regarding the open-source datasets integrated into our study, we strictly adhere to their respective licenses and terms of use.

\section*{Acknowledgment}
We thank all the anonymous reviewers for their insightful and valuable comments. This work is supported by the National Natural Science Foundation of China (Grant No. U21B2009, 62376027).

\bibliography{custom}

\newpage

\appendix

\section{Details of Data Construction}
\label{sec:appendix_detail_data}
In this section, we introduce more details of the data construction.

\begin{figure*}[t]
    \centering
    \hfill \hyperlink{back:pre_llm_filtering}{\textbf{[Back to Text]}}
    \includegraphics[width=0.98\linewidth]{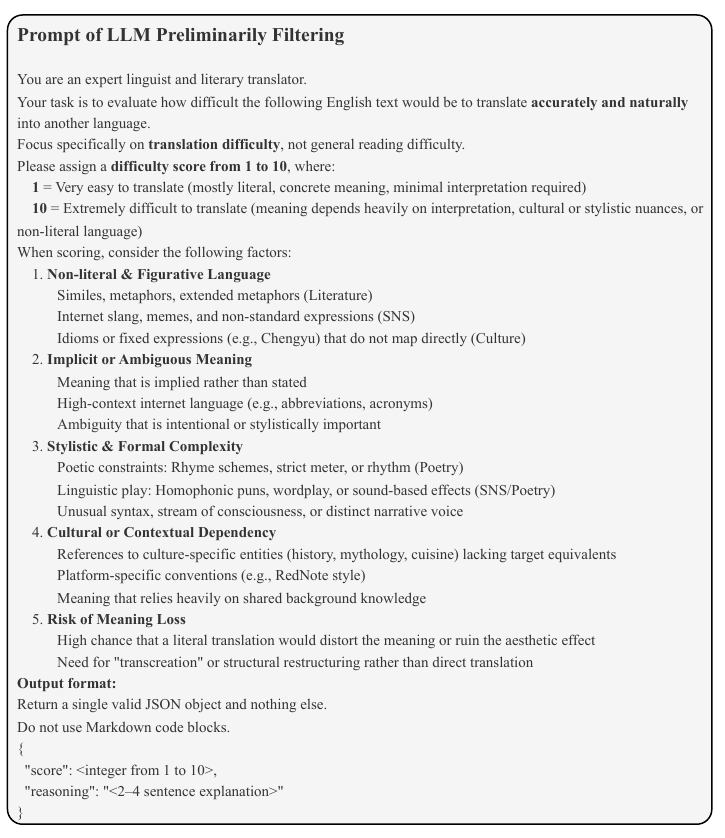}
    \caption{The prompt of LLM preliminary filtering before manual inspection.}
    \label{fig:pre_llm_filtering}
\end{figure*}

\hypertarget{back:pre_llm_filtering}{\subsection{Preliminarily Filtering with LLM}}
\label{sec:appendix_llm_filtering}
To ensure the curated MENT dataset maintains a high concentration of linguistically challenging and non-literal samples, we employ LLM as an initial automated filter.
As illustrated in Figure \ref{fig:pre_llm_filtering}, the LLM is prompted to assign higher scores to source sentences that exhibit challenges of non-literal translation, and the LLM is also required to generate a rationale for each score.
This explanatory output serves as assistance during the subsequent manual inspection phase, allowing human experts to quickly pinpoint the specific linguistic complexities, such as slang, cultural idiom.
Only the source sentences with higher score will go through the further manual inspection.

\hypertarget{back:data_annotation}{\subsection{Data Annotation}}
\label{sec:appendix_annotation}
To mitigate annotator fatigue and maintain consistency, we limit the daily workload of each annotator to approximately 100 ratings. This corresponds to evaluating the translation outputs of all 10 systems for 10 source sentences.
All annotators are recruited as interns and receive fair compensation at competitive market rates, and are clearly informed that the data they annotated would be used for academic research.

The detail instruction of human annotation on evaluating the translation quality is shown in Figure \ref{fig:annotation_criteria}.
The detail instruction of human annotation on reference is shown in Figure \ref{fig:annotation_reference}.

\begin{figure}[t]
    \centering
    \includegraphics[width=0.98\linewidth]{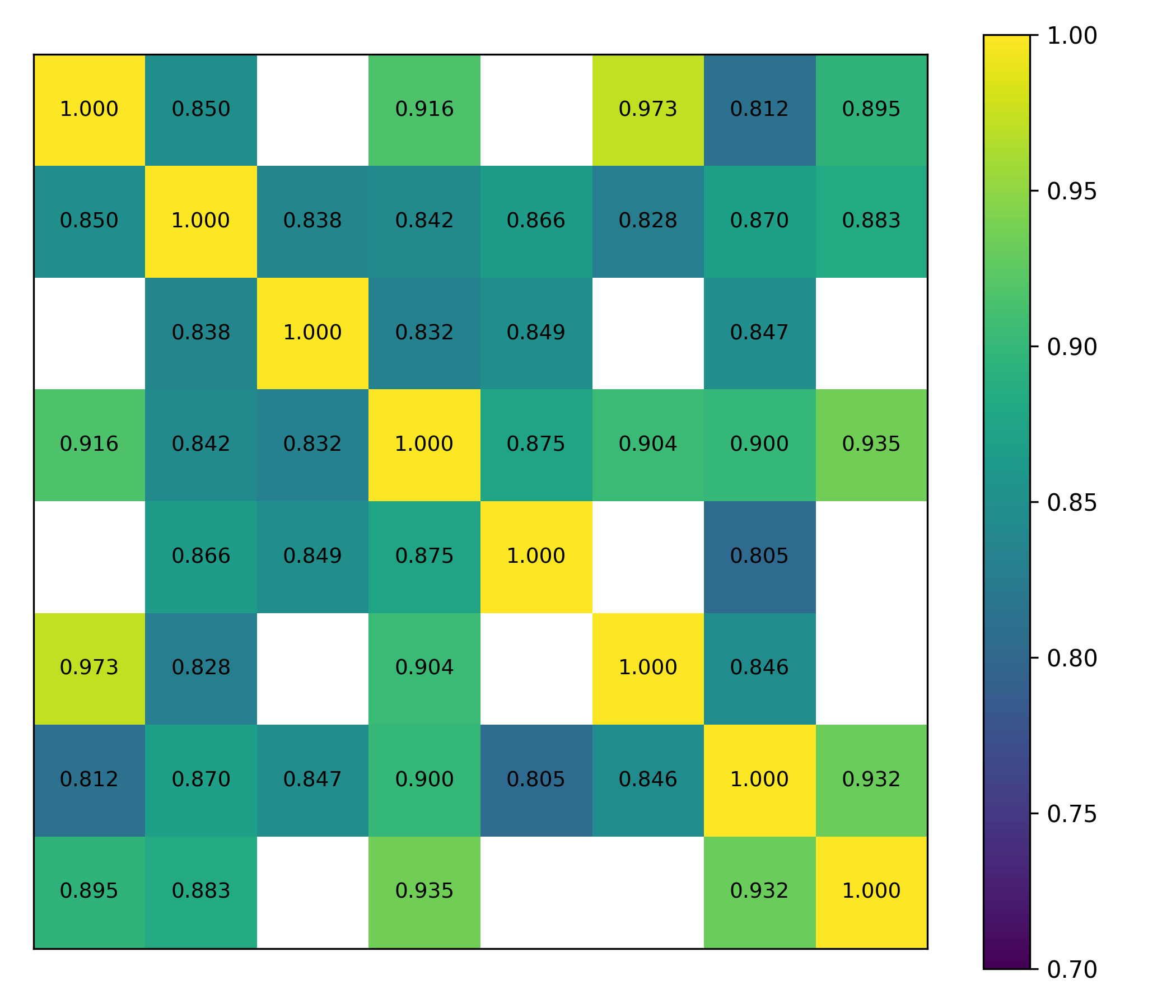}
    \caption{Heatmap of Pearson correlations for segment-level inter-annotator agreement. Each cell $(i, j)$ displays the average correlation coefficient for shared translation segments. Blank cells denote pairs with no overlapping annotation tasks due to the workload distribution. 
    }
    \label{fig:segment_annotator_correlation}
\end{figure}

\begin{figure*}[t]
\centering
\begin{tcolorbox}[
  colback=gray!5,
  colframe=black!40,
  title={5-Point SQM Human Annotation Criteria \hfill \hyperlink{back:data_annotation}{[Back to Text]}},
  fonttitle=\bfseries,
  left=6pt,right=6pt,top=4pt,bottom=4pt,
  width=\textwidth,
  boxrule=0.5pt
]
\textbf{Task Description}\\
Annotators are asked to evaluate the machine translation quality of source texts containing high-context linguistic features (e.g., slang, metaphors, ancient poetry), which are hard to understand and cannot be accurately translated through literal translation alone. 
Special attention should be paid to whether the translation captures the figurative meaning rather than just the literal surface form.\\
\textbf{Rating Scale}\\
The quality is assessed on a 5-point scale (0-4):\\
\quad \textbf{Score 0: Severe Knowledge Failure / Nonsense} The translation contains severe errors or omissions in understanding and translating the knowledge contained in the source text.\\
\quad \textbf{Score 1: Partial Severe Error} The translation contains severe errors or omissions in \textit{parts} of the knowledge understanding and translation.\\
\quad \textbf{Score 2: Comprehensible but Biased / Literal} The understanding and translation of the source knowledge have deviations (bias) or rely on literal translation, but the content remains generally understandable.\\
\quad \textbf{Score 3: Accurate but Unfluent} The understanding and translation of the source content are \textbf{entirely correct} (including slang/idioms), but the translation is not fluent or contains minor grammatical/register errors.\\
\quad \textbf{Score 4: Excellent / Culturally Adaptive} The understanding and translation of the source content are \textbf{entirely correct}, AND the expression is fluent and authentic.
\end{tcolorbox}
\caption{Human annotation criteria of translation quality. All recruited annotators hold degrees in translation and possess extensive prior experience in translation evaluation tasks. }
\label{fig:annotation_criteria}
\end{figure*}

\begin{figure*}[t]
\centering
\begin{tcolorbox}[
  colback=gray!5,
  colframe=black!40,
  title={Human Annotation of Translation Reference \hfill \hyperlink{back:data_annotation}{[Back to Text]}},
  fonttitle=\bfseries,
  left=6pt,right=6pt,top=4pt,bottom=4pt,
  width=\textwidth,
  boxrule=0.5pt
]
\textbf{Task Description}\\
Annotators are asked to curate the final reference translation based on the source text and five candidate translations.

\textbf{Annotation Steps}\\
Review the five candidate translations. Select the one with the highest translation quality.

Determine whether the selected translation requires modification. If yes, please modify it to the final version.
\end{tcolorbox}
\caption{Human annotation steps of translation reference.}
\label{fig:annotation_reference}
\end{figure*}

\subsection{Calculation of IAA}
\label{sec:appendix_IAA}
The IAA for each pair of annotators is determined as follows:

Let $\mathcal{K}_{i,j}$ denote the set of segments commonly annotated by annotators $i$ and $j$.
For a segment $k \in \mathcal{K}_{i,j}$, let $\mathbf{v}_{i,k} \in \mathbb{N}^{10}$ represent the score vector assigned by annotator $i$ to the 10 MT systems. The correlation $r_{i,j}$ is defined as:
\begin{equation}
    r_{i,j} = \operatorname{Pearson}\left( \sum_{k \in \mathcal{K}_{i,j}} \mathbf{v}_{i,k}, \quad \sum_{k \in \mathcal{K}_{i,j}} \mathbf{v}_{j,k} \right)
\end{equation}
\noindent where the summation $\sum$ is performed element-wise over the vectors. The resulting correlation measures the agreement between annotators regarding the relative quality of the 10 MT systems.

The visualization of segment-level IAA is shown in Figure \ref{fig:segment_annotator_correlation}.

\begin{figure*}[t]
    \centering
    \hfill \hyperlink{back:data_sample}{\textbf{[Back to Text]}}
    \includegraphics[width=0.98\linewidth]{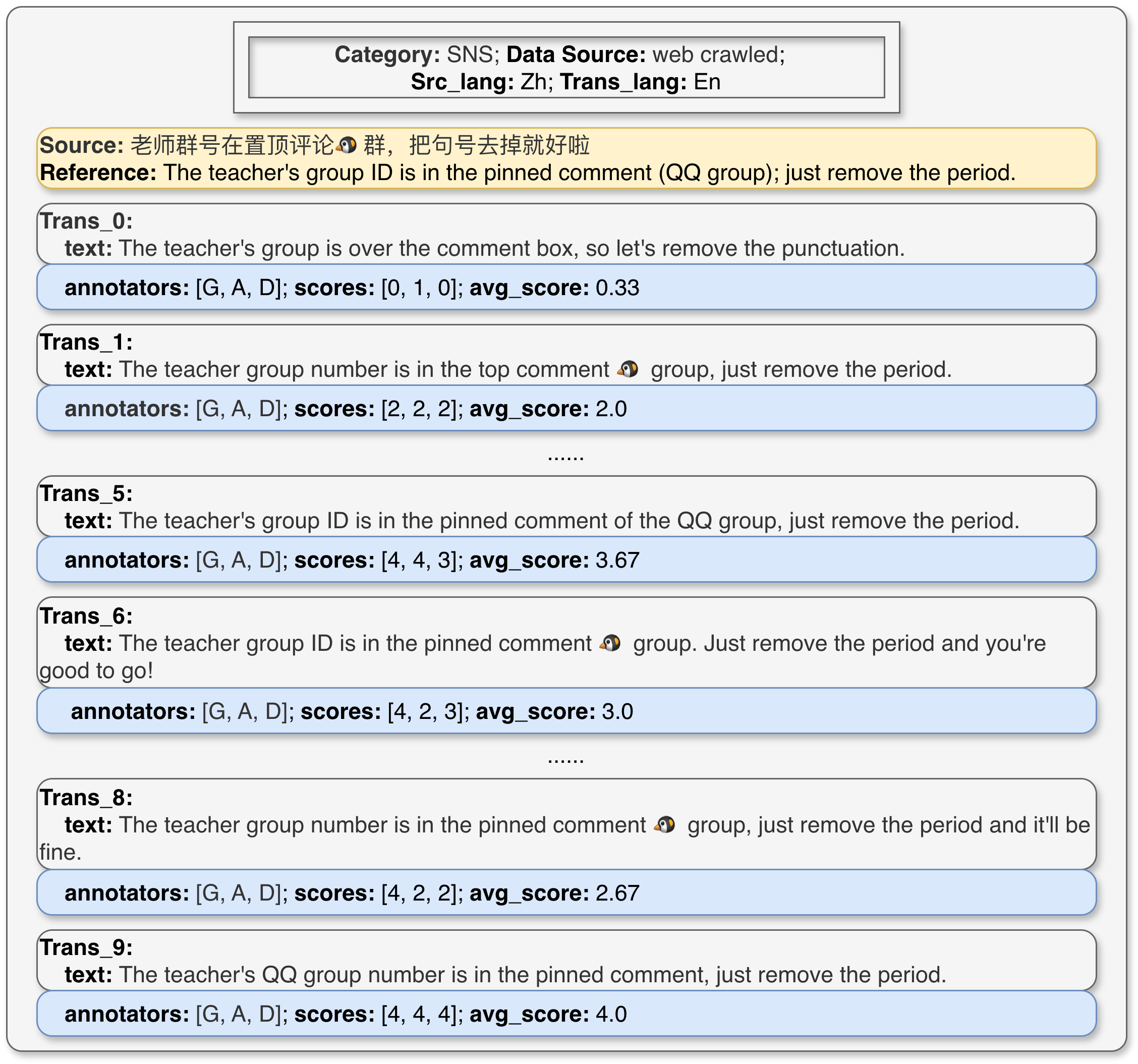}
    \caption{Sample (SNS domain, Zh-En) from MENT, the annotated data comprises a reference and scores of translation quality.}
    \label{fig:dataset_sample_1}
\end{figure*}

\begin{figure*}[t]
    \centering
    \hfill \hyperlink{back:data_sample}{\textbf{[Back to Text]}}
    \includegraphics[width=0.98\linewidth]{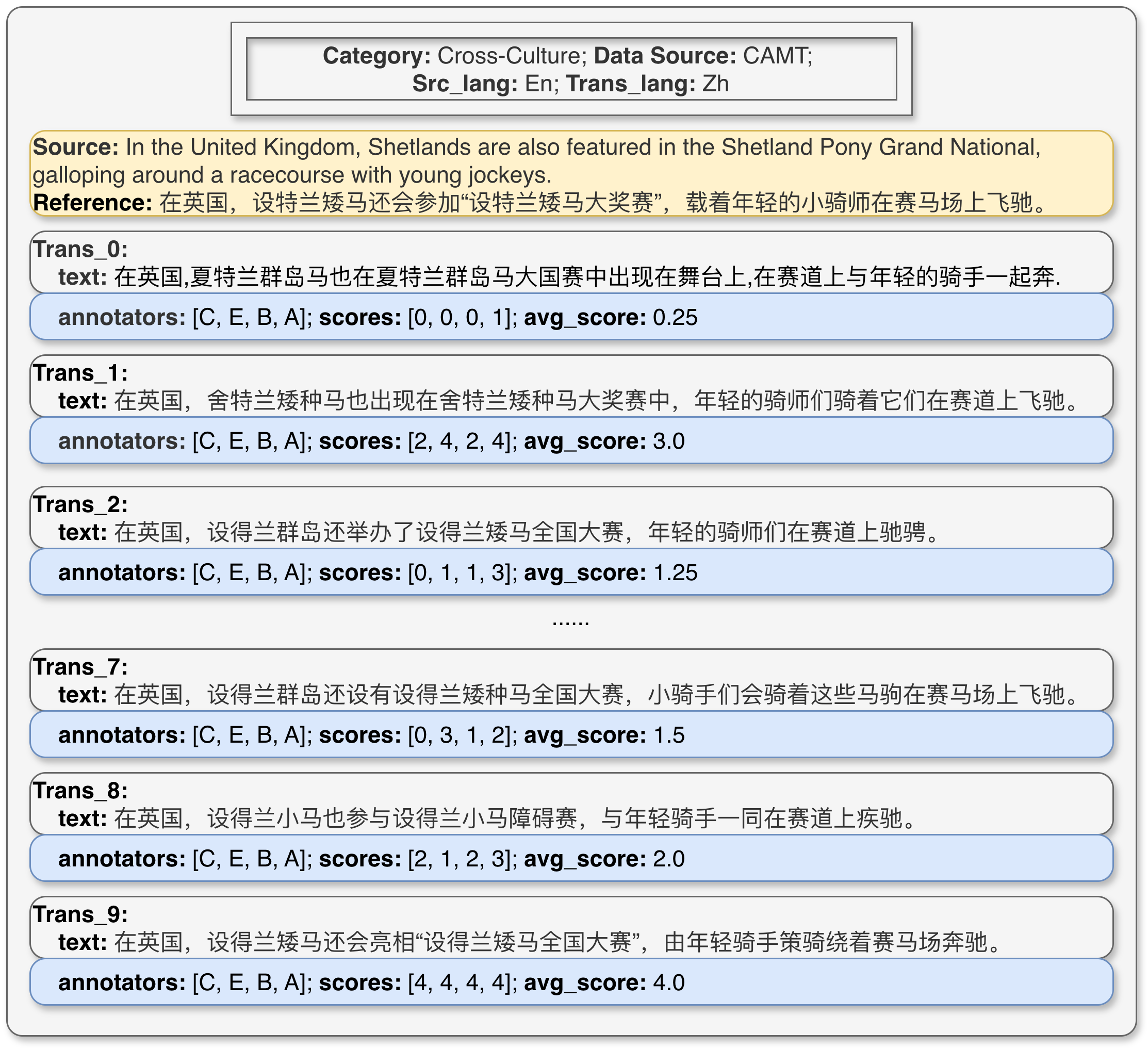}
    \caption{Sample (Cross-Culture domain, En-Zh) from MENT, the annotated data comprises a reference and scores of translation quality.}
    \label{fig:dataset_sample_2}
\end{figure*}

\begin{figure*}[t]
    \centering
    \hfill \hyperlink{back:data_sample}{\textbf{[Back to Text]}}
    \includegraphics[width=0.98\linewidth]{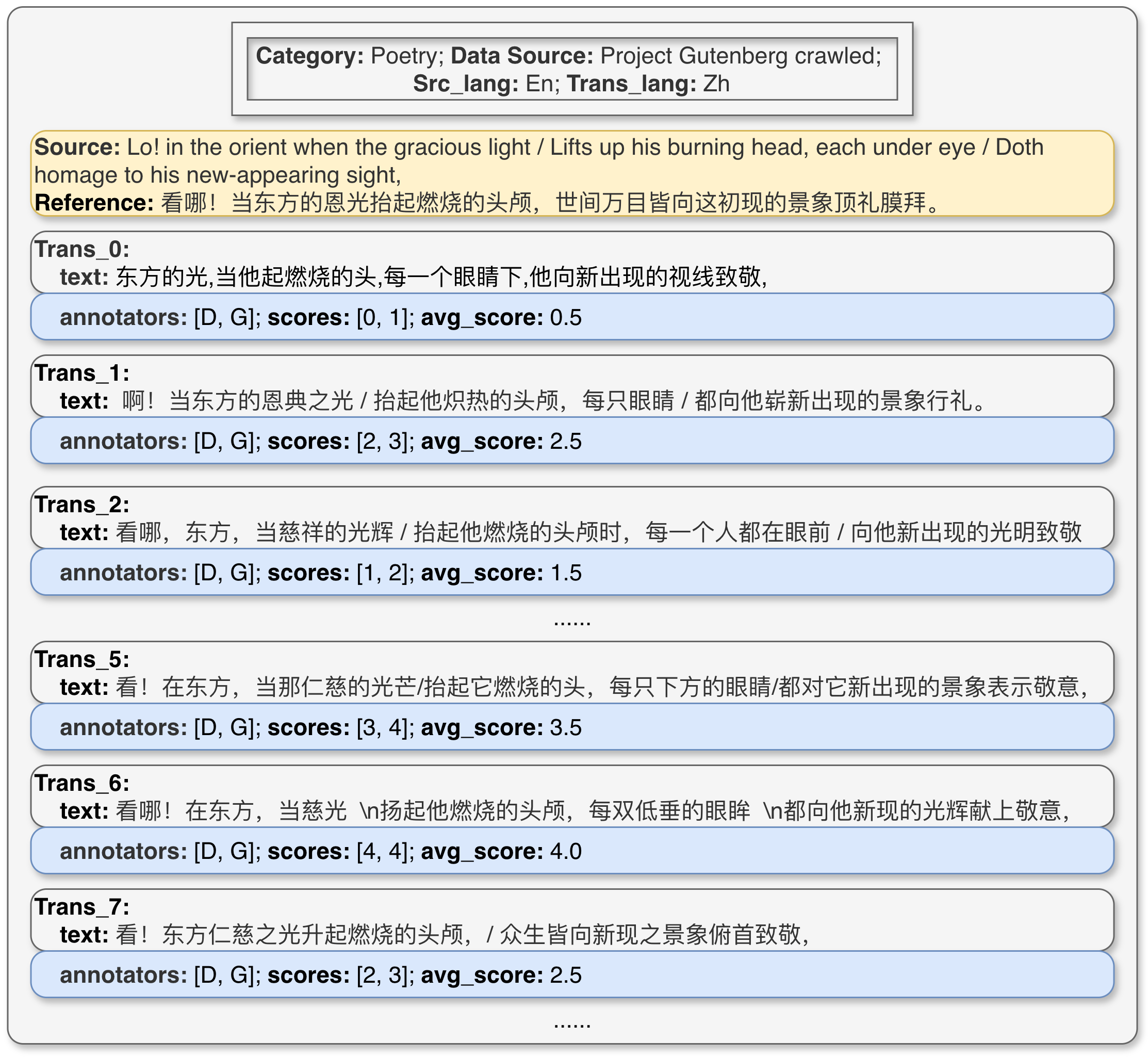}
    \caption{Sample (Poetry domain, En-Zh) from MENT, the annotated data comprises a reference and scores of translation quality.}
    \label{fig:dataset_sample_3}
\end{figure*}

\begin{figure*}[t]
    \centering
    \hfill \hyperlink{back:data_sample}{\textbf{[Back to Text]}}
    \includegraphics[width=0.98\linewidth]{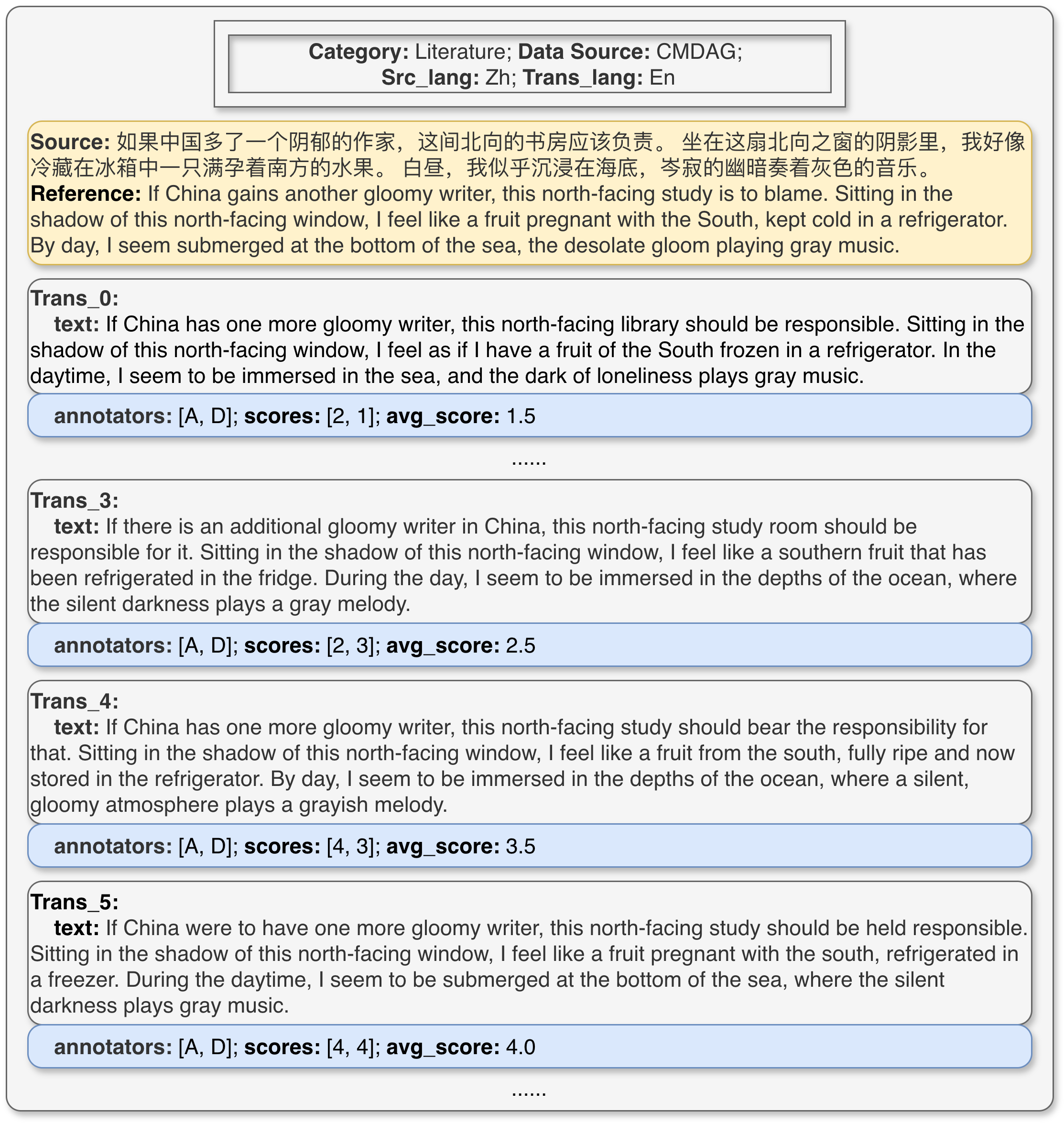}
    \caption{Sample (Literature domain, Zh-En) from MENT, the annotated data comprises a reference and scores of translation quality.}
    \label{fig:dataset_sample_4}
\end{figure*}

\hypertarget{back:data_sample}{\subsection{Samples from Dataset}}
\label{sec:appendix_data_sample}
We illustrate samples from MENT covering each domain, as shown in Figure \ref{fig:dataset_sample_1}, \ref{fig:dataset_sample_2}, \ref{fig:dataset_sample_3}, and \ref{fig:dataset_sample_4}.

\section{Implementation of RATE}
\label{sec:appendix_rate_impl}
In this section, we introduce the detailed implementation of each agent in RATE.

\begin{figure*}[t]
    \centering
    \hfill \hyperlink{back:core_agent}{\textbf{[Back to Text]}}
    \includegraphics[width=0.98\linewidth]{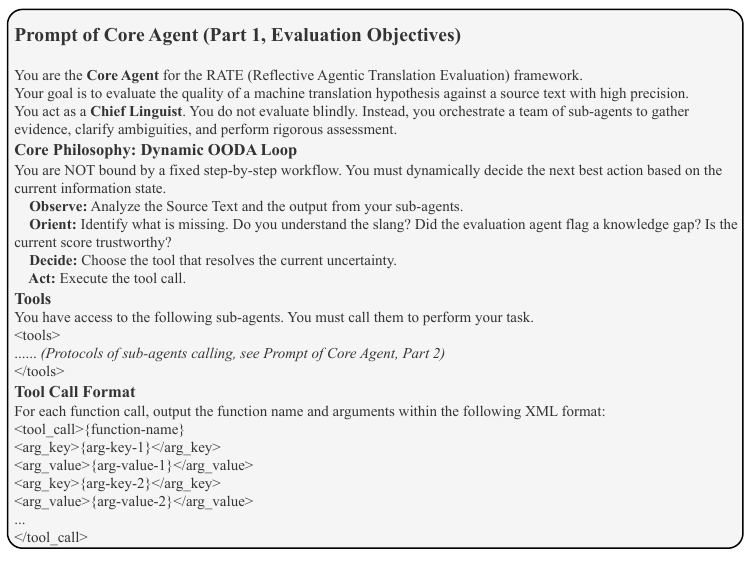}
    \caption{Prompt of Core Agent (part 1), outlining the evaluation objectives.}
    \label{fig:core_agent_prompt_1}
\end{figure*}

\begin{figure*}[t]
    \centering
    \hfill \hyperlink{back:core_agent}{\textbf{[Back to Text]}}
    \includegraphics[width=0.98\linewidth]{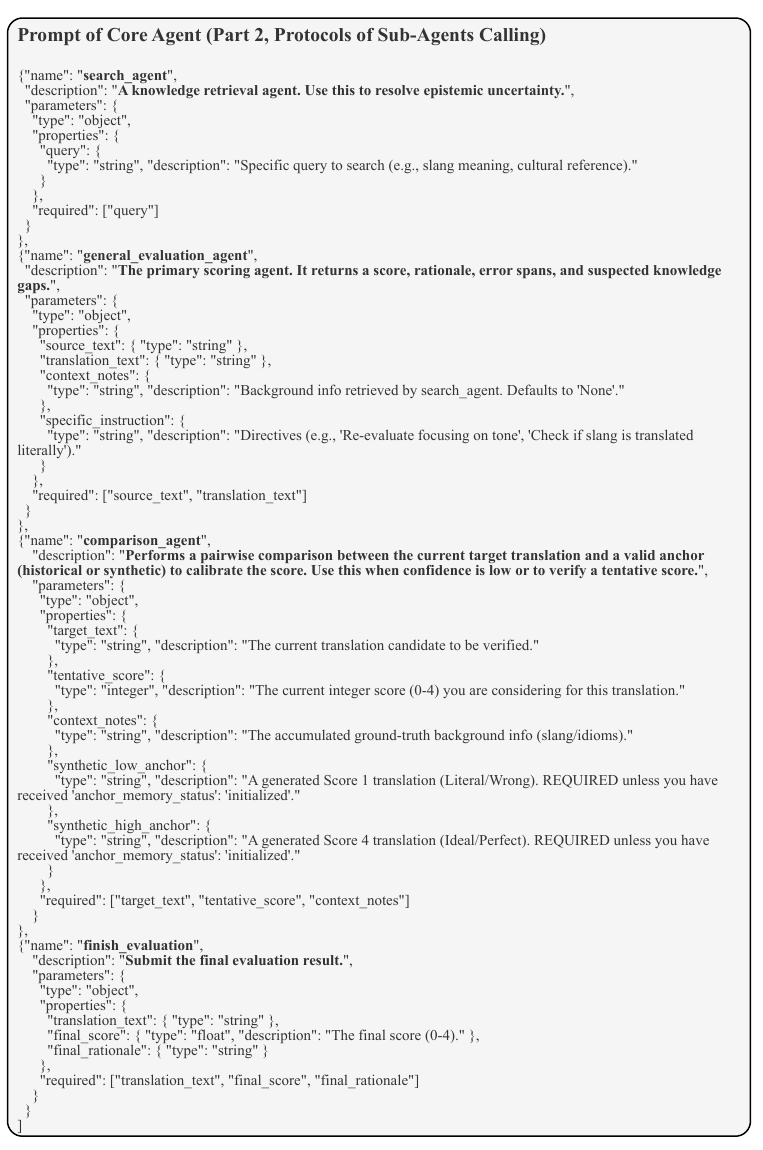}
    \caption{Prompt of Core Agent (part 2), outlining protocols of sub-agents calling.}
    \label{fig:core_agent_prompt_2}
\end{figure*}

\begin{figure*}[t]
    \centering
    \hfill \hyperlink{back:core_agent}{\textbf{[Back to Text]}}
    \includegraphics[width=0.98\linewidth]{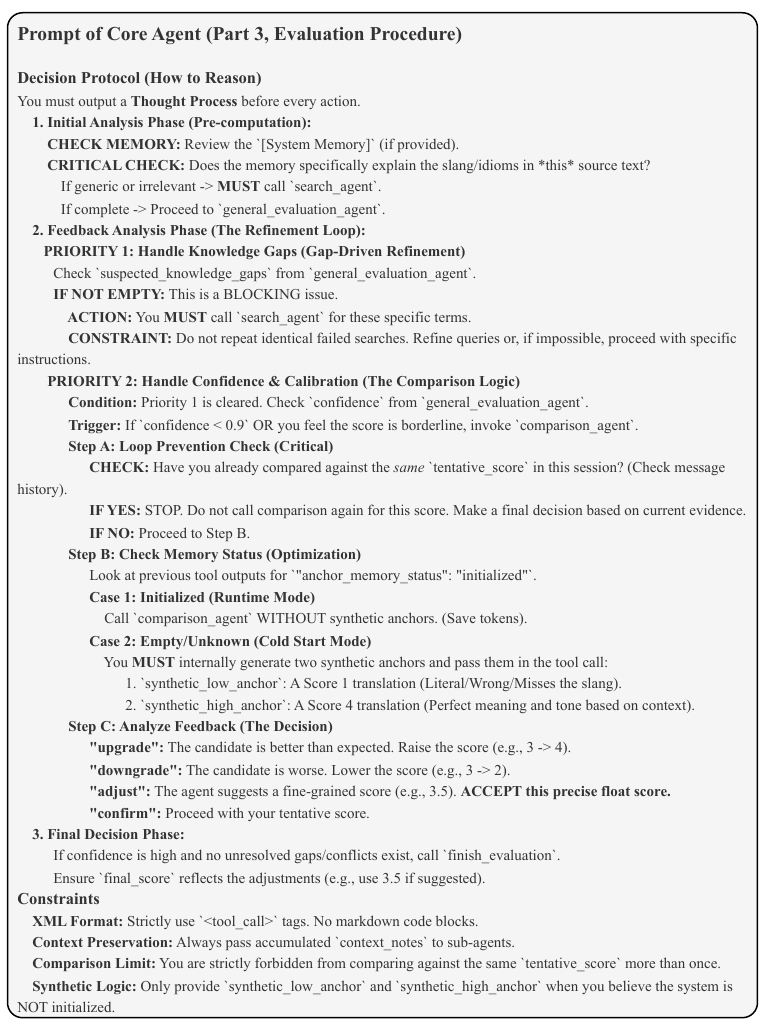}
    \caption{Prompt of Core Agent (part 3), outlining the evaluation procedure.}
    \label{fig:core_agent_prompt_3}
\end{figure*}

\hypertarget{back:core_agent}{\paragraph{Core Agent.}} The Core Agent is powered by \texttt{GLM-4.6}, one of the best performance large-scale foundation models on agentic tasks.
The prompt for Core Agent is shown in Figure \ref{fig:core_agent_prompt_1}, \ref{fig:core_agent_prompt_2}, and \ref{fig:core_agent_prompt_3}.
To optimize efficiency during the evaluation of a specific source sentence, the Core Agent accumulates and persists the knowledge retrieved by the Search Agent in the context memory. This mechanism ensures that external queries for identical background information are not redundantly invoked.
Furthermore, to prevent potential infinite loops, we implement a maximum loop threshold $t$ ($t=10$). If the reflective loop reaches this limit, the Core Agent is mandated to immediately output the final evaluation result.

\begin{figure*}[t]
    \centering
    \hfill \hyperlink{back:eval_agent}{\textbf{[Back to Text]}}
    \includegraphics[width=0.98\linewidth]{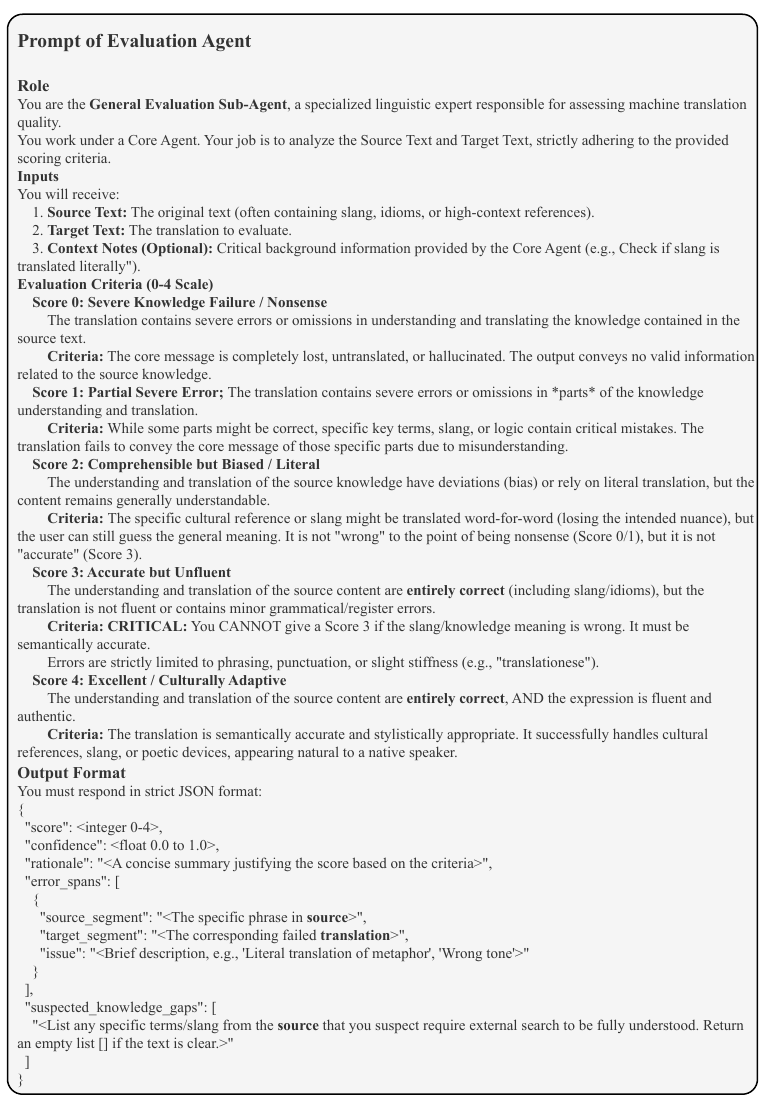}
    \caption{Prompt of Evaluation Agent.}
    \label{fig:eval_agent_prompt}
\end{figure*}

\hypertarget{back:eval_agent}{\paragraph{Evaluation Agent.}} The Evaluation Agent is build upon \texttt{GPT-4o} and is tasked with conducting translation quality assessment based on the instruction and context notes including necessary knowledge, provided by the Core Agent.
The structured output of this agent including numerical score, confidence, and identified error spans. Crucially, the agent is also responsible for flagging suspected knowledge gaps, such as slang or cultural idiom which is critical for evaluation. This feedback allows the Core Agent to locate the missing background information.
The prompt for Evaluation Agent is illustrated in Figure \ref{fig:eval_agent_prompt}.

\begin{figure*}[t]
    \centering
    \hfill \hyperlink{back:search_agent}{\textbf{[Back to Text]}}
    \includegraphics[width=0.98\linewidth]{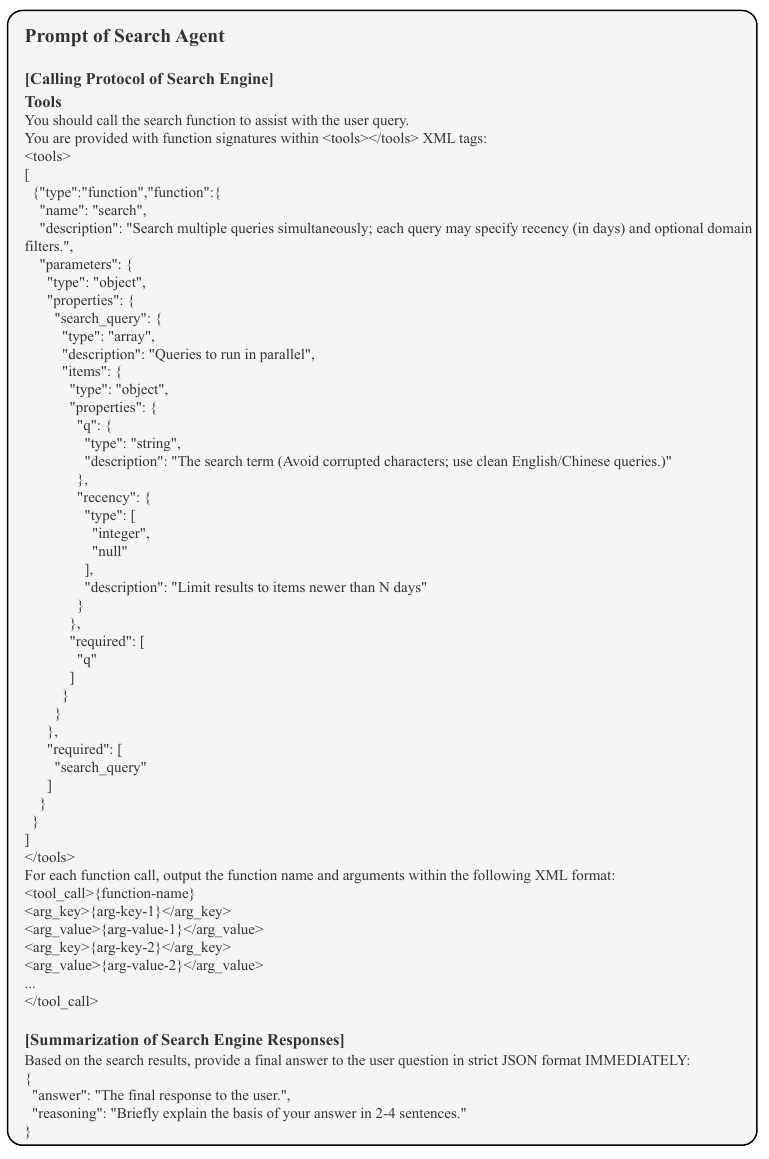}
    \caption{Prompt of Search Agent, including the calling protocol of search engine, and the summarization of search engine responses.}
    \label{fig:search_agent_prompt}
\end{figure*}

\hypertarget{back:search_agent}{\paragraph{Search Agent.}} The backbone model for Search Agent is \texttt{GLM-4.6}, selected for its requirement of tool-calling precision and reliability in executing structured function calls.
Acting as the framework's gateway to real-time information, this agent is designed to resolve specific queries of Core Agent.
The Search Agent first analyzes the request from Core Agent and reformulates it into search queries. 
By parsing the structured tool calls, we use \texttt{Bing} search engine to retrieve real-time external information.
Once the raw search results are retrieved, the Search Agent summarizes the most relevant information along with a brief explanation, and returns it back into Core Agent.
The prompt for Search Agent is shown in Figure \ref{fig:search_agent_prompt}, including both search engine calling protocol, and the summarization of search responses.

\hypertarget{back:compare_agent}{\paragraph{Comparison Agent.}} The Comparison Agent is based on \texttt{GPT-4o}, aiming to calibrate the pointwise score provided by Evaluation Agent.
Rather than assessing a translation in isolation, the Comparison Agent performs a pairwise comparison between the current translation and established anchors.
The agent maintains a dynamic memory of translation anchors, utilizing previously evaluated translations.
Recognizing that this memory is initially empty (Cold Start Mode), we design a bootstrapping strategy: when the Core Agent first invokes the Comparison Agent, it should provide two synthetic anchors, a Score 1 anchor (a poor, literal translation), and a Score 4 anchor (a high-quality, context-aware translation based on the retrieved background knowledge) for initialization. 
After the memory is initialized, the system transitions into Runtime Mode, and the Comparison Agent identifies the anchor with the closest score to the current evaluating translation for pairwise comparison. 
Once the Core Agent finalizes a score for a translation, this translation is then updated into the corresponding score-level anchor slot, iteratively refining the calibration pool for subsequent evaluation of the same source sentence.

To mitigate the position bias in pairwise comparisons, we adopt a bidirectional evaluation strategy \citep{2024-faireval}.
For each assessment, the translation and the anchor are swapped their position and evaluated twice. Each individual comparison yields one of three outcomes for the translation: \textbf{Win} (superior to anchor), \textbf{Tie} (equivalent to anchor), or \textbf{Lose} (inferior to anchor).
The combination of these twice comparison results leads to five distinct calibration scenarios, which the Comparison Agent uses to refine the tentative score:
\textbf{(1) Win-Win:} The translation consistently better than the anchor. If the current score is not higher than the anchor's, an upgrade of 1.0 is recommended; if the current score is higher than the anchor's, the current score is maintained.
\textbf{(2) Win-Tie:} The translation is borderline better than the anchor. If the current score is not higher than the anchor's, an fine-grained upgrade of 0.5 is recommended; if the current score is higher than the anchor's, the current score is maintained.
\textbf{(3) Tie-Tie or Win-Lose:} The translation is equivalent to the anchor, and the current score is maintained.
\textbf{(4) Lose-Tie:} The translation is borderline worse than the anchor, and a downgrade of 0.5 is recommended.
\textbf{(5) Lose-Lose:} The translation is consistently worse than the anchor, and a downgrade of 1.0 is recommended.

\begin{figure*}[t]
    \centering
    \hfill \hyperlink{back:compare_agent}{\textbf{[Back to Text]}}
    \includegraphics[width=0.98\linewidth]{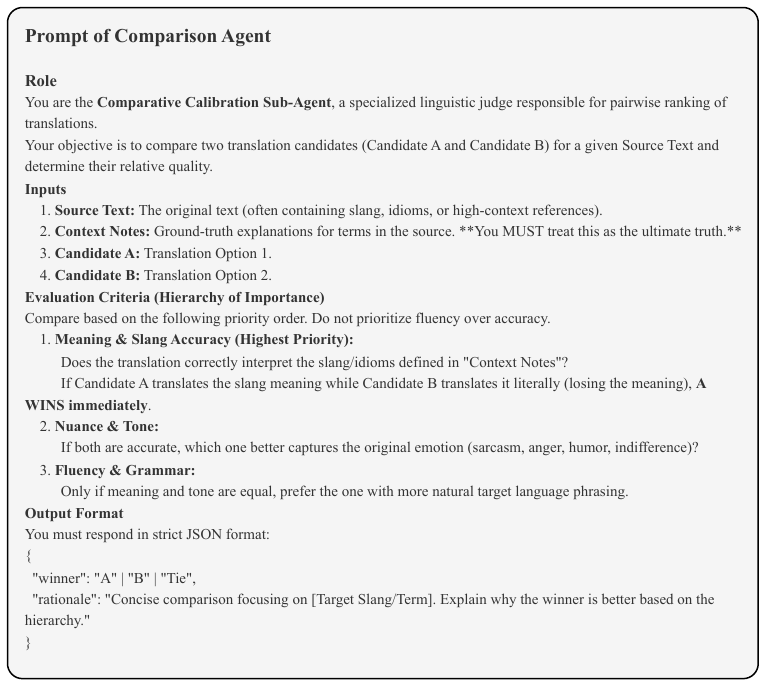}
    \caption{Prompt of Comparison Agent.}
    \label{fig:comparison_agent_prompt}
\end{figure*}

The prompt of Comparison Agent is shown in Figure \ref{fig:comparison_agent_prompt}.

\begin{figure*}[t]
    \centering
    \hfill \hyperlink{back:rate_trajectory}{\textbf{[Back to Text]}}
    \includegraphics[width=0.98\linewidth]{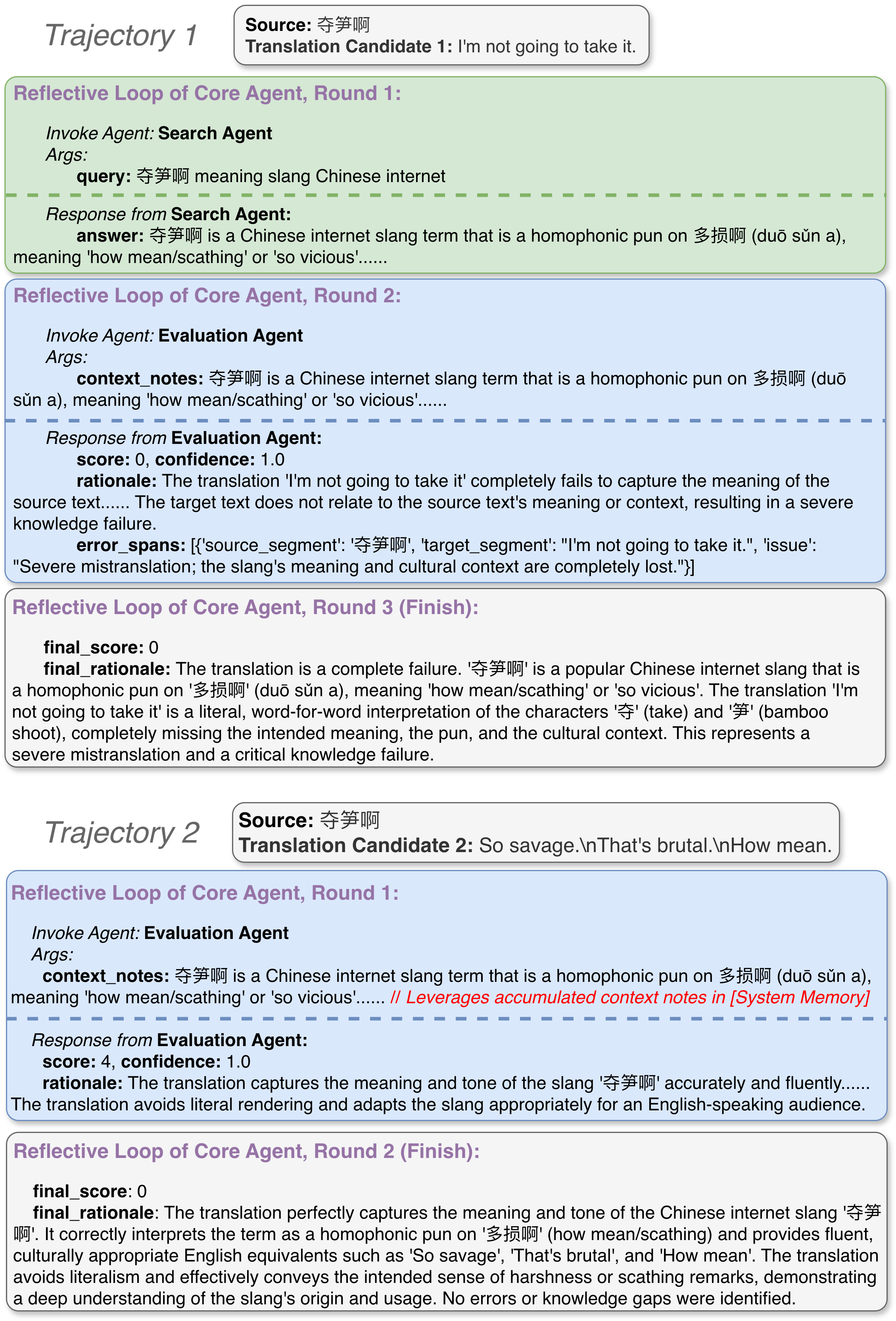}
    \caption{Trajectory of RATE, illustrating the invoking of Search Agent to retrieve background knowledge, and instructing the Evaluation Agent to assign a score with the knowledge.}
    \label{fig:rate_trajectory_1}
\end{figure*}

\begin{figure*}[t]
    \centering
    \hfill \hyperlink{back:rate_trajectory}{\textbf{[Back to Text]}}
    \includegraphics[width=0.98\linewidth]{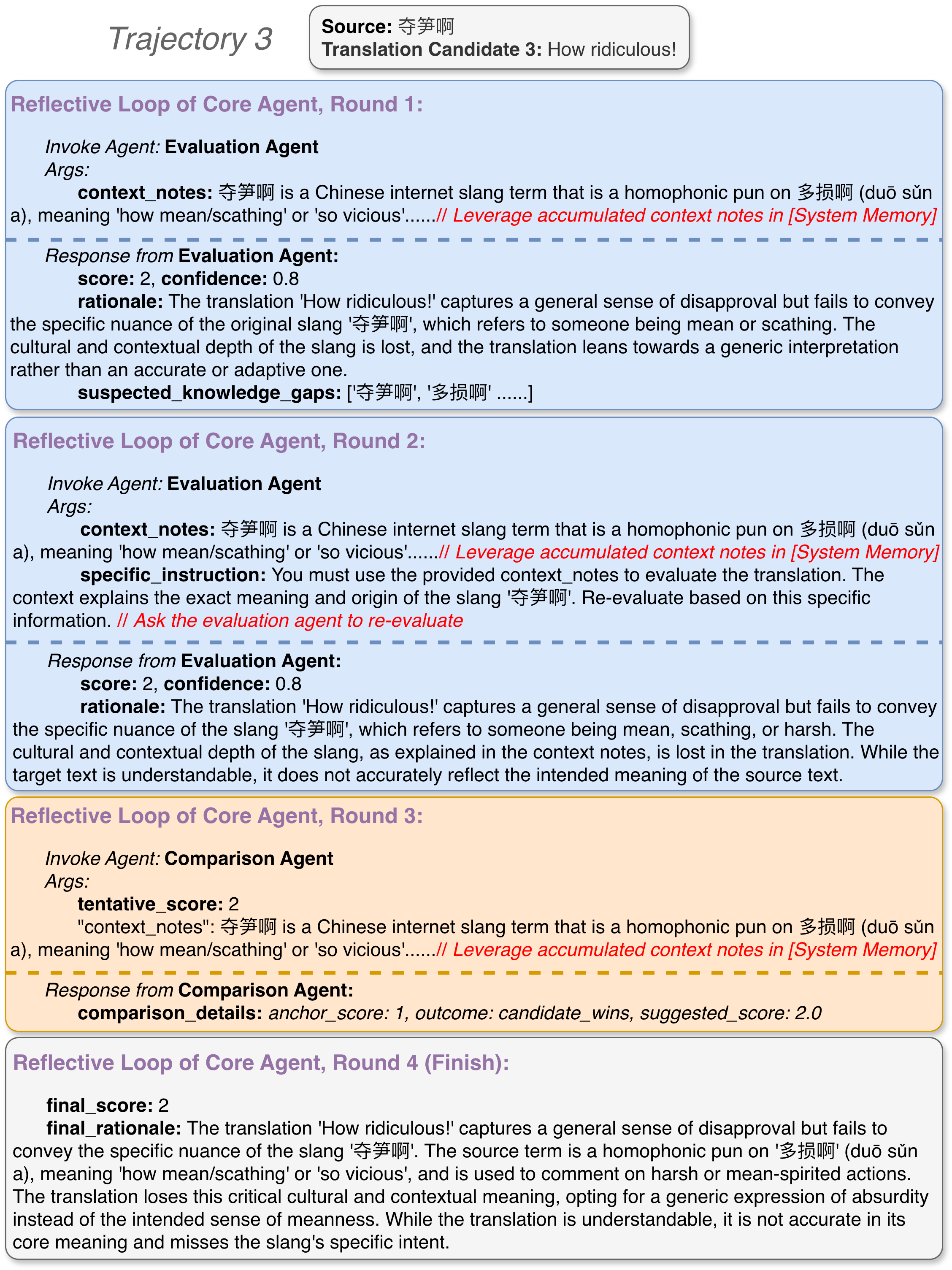}
    \caption{Trajectory of RATE, illustrating the Evaluation Agent fails to reach high confidence despite specific instructions. The Core Agent dynamic switches strategy by invoking the Comparison Agent, and successfully calibrates the score through pairwise comparison.}
    \label{fig:rate_trajectory_2}
\end{figure*}

\hypertarget{back:rate_trajectory}{\section{Evaluation Trajectories of RATE}}
\label{sec:appendix_rate_sample}
In this section, we illustrate evaluation trajectories of RATE\footnote{We will release all evaluation trajectories of RATE as soon as possible.}, as shown in Figure \ref{fig:rate_trajectory_1}, and \ref{fig:rate_trajectory_2}.

\paragraph{Trajectory 1}
In the evaluation of Translation Candidate 1 (Figure \ref{fig:rate_trajectory_1}), observing the unknown slang, the Core Agent firstly invokes Search Agent.
The agent successfully retrieves and summarizes for Chinese homophonic pun from search engine.
Subsequently, the Core Agent integrates the retrieved knowledge into the context notes to guide the Evaluation Agent.
Equipped with the background information, the Evaluation Agent correctly identifies the Translation Candidate 1 as a literally word-to-world translation, assigning it a score of 0.
After receiving the high-confidence assessment from Evaluation Agent, the Core Agent then terminates the reflective loop and outputs the final score.

\paragraph{Trajectory 2}
In the evaluation of Translation Candidate 2 (Figure \ref{fig:rate_trajectory_1}), instead of invoking a new Search Agent, the Core Agent leverages the background knowledge accumulated from the previous trajectory that accumulated in [System Memory].
By directly instructing the Evaluation Agent with the accumulated knowledge, the Evaluation Agent recognizes that Candidate 2 correctly interprets the homophonic pun and offers a fluent and culturally appropriate translation, resulting in a high score of 4.
The Core Agent then concludes the evaluation based on the high-confidence judgment from Evaluation Agent.

\paragraph{Trajectory 3}
In the evaluation of Translation Candidate 3 (Figure \ref{fig:rate_trajectory_2}), the Core Agent initially invokes the Evaluation Agent with the accumulated context which is similar to Trajectory 2.
However, the candidate (``How ridiculous!'') is neither a complete mistranslation nor fully accurate, causing the Evaluation Agent to flag a suspected knowledge gap and request clarification.

Although the Core Agent re-invokes the Evaluation Agent with specific instructions confirming the context, the agent maintains its score with sub-optimal confidence.
To resolve this, the Core Agent dynamically switches strategy by selecting the Comparison Agent for pairwise calibration. 
The comparison reveals that the candidate is superior to the low-quality anchor of score 1, acknowledging that while the translation captures the general sense, it still fails to convey the specific nuance of the original slang, and the Core Agent assigns the candidate of score 2.

\section{Experiment on Different Judge Models}
We conduct an additional experiment to evaluate the impact of different selection judge models. 
As introduced in Appendix \ref{sec:appendix_rate_impl}, in our main settings, the Core Agent of RATE is built upon \texttt{GLM-4.6}, while the Evaluation Agent used to judge the translation quality is based on \texttt{GPT-4o}.

To investigate the generalizability of our framework across different backbones, we design an experiment where the backbone models for all agents within RATE are switched into \texttt{GLM-4.6}. 
We select \texttt{GEMBA-DA} as the baseline for comparison, as it demonstrates the strongest performance among all LLM-as-a-Judge baselines in our main experiments (Table \ref{tab:main_results}).

\begin{table}[ht]
\centering
\small
    \begin{tabular}{l ccc}
    \toprule
    \multirow{2}{*}{\textbf{Metric}} & \multicolumn{3}{c}{\textbf{Meta}}  \\
    \cmidrule(lr){2-4}
    & ZH-EN & EN-ZH & Avg. \\
    \midrule
    \rowcolor{gray!15}
    \multicolumn{4}{l}{\textit{Backbone: GPT-4o}} \\
    GEMBA-DA & 76.7 & 77.7 & 77.2 \\
    RATE (Ours) & 83.3 & 77.4 & 80.4 \\
    \midrule
    \rowcolor{gray!15}
    \multicolumn{4}{l}{\textit{Backbone: GLM-4.6}} \\
    GEMBA-DA & 78.1 \color{green!70!black}{(+1.4)} & 77.0 \color{red}{(-0.7)} & 77.6 \color{green!70!black}{(+0.4)} \\
    RATE (Ours) & 82.7 \color{red}{(-0.6)} & 79.4 \color{green!70!black}{(+2.0)} & 81.1 \color{green!70!black}{(+0.7)} \\
    \bottomrule
    \end{tabular}
\caption{Comparison of RATE against the best performance LLM-as-a-Judge baseline (GEMBA-DA) using different backbone LLMs (GPT-4o vs. GLM-4.6). The values in parentheses denote the performance gap compared to the GPT-4o backbone, highlighting the minimal variance and robustness of RATE.}
\label{tab:switch_judger}
\end{table}

The experimental results are presented in Table \ref{tab:switch_judger}. As shown in the table, replacing the backbone model yields a slight performance improvement for both methods on the average meta scores.
Crucially, RATE consistently maintains superior performance over the baseline, confirming that our framework's effectiveness holds across different backbone models.

\section{System-level SPA on MENT}
\label{sec:appendix_spa}
To further validate the robustness of evaluation metrics in discerning system-level translation quality differences, we report the Soft Pairwise Accuracy (SPA) \citep{2024-spa} in Table \ref{tab:spa_results}, complementing the correlation analysis in Table \ref{tab:main_results}.
As illustrated in Table \ref{tab:spa_results}, our RATE achieves a near-perfect SPA on ZH-EN and maintains competitive performance on EN-ZH, further demonstrating its effectiveness in providing reliable system-level rankings for non-literal translations.

\section{Details of Experimental Setup}
\label{sec:appendix_exp_setup}
In this section, we provide a comprehensive description of the evaluation MT metrics investigated in our study.

\subsection{Meta-Evaluation}
Complementing the standard meta-evaluation statistics from WMT23 Metrics Shared Task \citep{2023-wmtmetrics}, we incorporate Spearman correlation to provide a more robust assessment of rank-order consistency. 
The final meta score is derived by equally weighting the following statistics across both translation directions.
\begin{itemize}
    \item System-level pairwise ranking accuracy (Acc) \citep{2021-acc}.
    \item Segment-level pairwise ranking accuracy with tie calibration (Acc-t) \citep{2023-acc-t}.
    \item System- and segment-level Pearson correlation ($r$).
    \item System- and segment-level Spearman correlation ($\rho$).
\end{itemize}

\begin{table}[t]
\centering
\small
\begin{tabular}{l cc}
\toprule
\textbf{Metric} & \textbf{ZH-EN} & \textbf{EN-ZH} \\
\midrule
\rowcolor{gray!15}
\multicolumn{3}{l}{\textit{Reference-based Metrics}} \\
BLEU         & 91.6 & 95.8 \\
BLEURT       & 94.7 & 79.9 \\
COMET        & 94.7 & 91.6 \\
XCOMET       & 90.5 & 85.2 \\
MetricX-23   & 87.3 & 83.1 \\
MetricX-24   & 87.3 & 84.2 \\
\midrule
\rowcolor{gray!15}
\multicolumn{3}{l}{\textit{Reference-free (QE) Metrics}} \\
COMETKiwi    & 76.8 & 75.7 \\
MetricX-23-QE & 74.7 & 69.4 \\
MetricX-24-QE & 75.7 & 78.9 \\
\midrule
\rowcolor{gray!15}
\multicolumn{3}{l}{\textit{LLM-as-a-Judge}} \\
GEMBA-MQM    & 91.6 & 95.8 \\
GEMBA-DA     & 93.7 & 95.8 \\
EAPrompt     & 91.6 & 88.4 \\
ThinMQM      & 90.5 & 95.8 \\
M-MAD        & 92.6 & 94.7 \\
\textbf{RATE (Ours)} & 100.0 & 94.7 \\
\bottomrule
\end{tabular}
\caption{System-level Soft Pairwise Accuracy (SPA) on ZH-EN and EN-ZH directions.}
\label{tab:spa_results}
\end{table}

\subsection{Evaluated Metrics}
Based on the characteristic of MT metrics, we categorize them into three paradigms and evaluated by MTME\footnote{\url{https://github.com/google-research/mt-metrics-eval}}, the standard metric evaluation tool recommended by WMT.
All evaluated metrics in the main experiment (Table \ref{tab:main_results}) are introduced as follows.
\subsubsection{Reference-based Metrics}
\paragraph{BLEU \citep{2002-bleu}:} We use SacreBLEU\footnote{\url{https://github.com/mjpost/sacrebleu}} to calculate BLEU. Score of each segment (a sentence) is calculated by sentence\_score(), while for each system is calculated by corpus\_score().

\paragraph{BLEURT \citep{2020-bleurt}:} We adopt the official evaluation scripts to calculate BLEURT score\footnote{\url{https://github.com/google-research/bleurt}}.

\paragraph{COMET \citep{2022-comet22}:} COMET score is calculated by \texttt{wmt22-comet-da}\footnote{\url{https://huggingface.co/Unbabel/wmt22-comet-da}}.

\paragraph{XCOMET \citep{2024-xcomet}:} XCOMET score is calculated by \texttt{XCOMET-XL}\footnote{\url{https://huggingface.co/Unbabel/XCOMET-XL}}.

\paragraph{MetricX \citep{2023-metricx, 2024-metricx}:} We evaluate both MetricX23 (\texttt{metricx-23-xl-v2p0}\footnote{\url{https://huggingface.co/google/metricx-23-xl-v2p0}}), and MetricX24 (\texttt{metricx-24-hybrid-xl-v2p6}\footnote{\url{https://huggingface.co/google/metricx-24-hybrid-xl-v2p6}}). 
For all MetricX series models, we adopt the official evaluation scripts\footnote{\url{https://github.com/google-research/metricx}}.

\subsubsection{Reference-free Metrics}
\paragraph{COMETKiwi \citep{2022-cometkiwi}:} COMETKiwi score is calculated by \texttt{wmt23-cometkiwi-da-xl}\footnote{\url{https://huggingface.co/Unbabel/wmt23-cometkiwi-da-xl}}.

\paragraph{MetricX-QE \citep{2023-metricx, 2024-metricx}:} We evaluate both MetricX23-QE (\texttt{metricx-23-qe-xl-v2p0}\footnote{\url{https://huggingface.co/google/metricx-23-qe-xl-v2p0}}), and MetricX24 (\texttt{metricx-24-hybrid-xl-v2p6}\footnote{\url{https://huggingface.co/google/metricx-24-hybrid-xl-v2p6}}), without providing references. 

\subsubsection{LLM-as-a-Judge}

Following \citet{2025-mad}, all LLM-as-a-Judge methods are configured with temperature=0 to ensure reproducibility, except where specifically noted.
All LLM-as-a-Judge methods adopt the reference-free paradigm.

\begin{figure*}[t]
    \centering
    \hfill \hyperlink{back:gemba_mqm}{\textbf{[Back to Text]}}
    \includegraphics[width=0.98\linewidth]{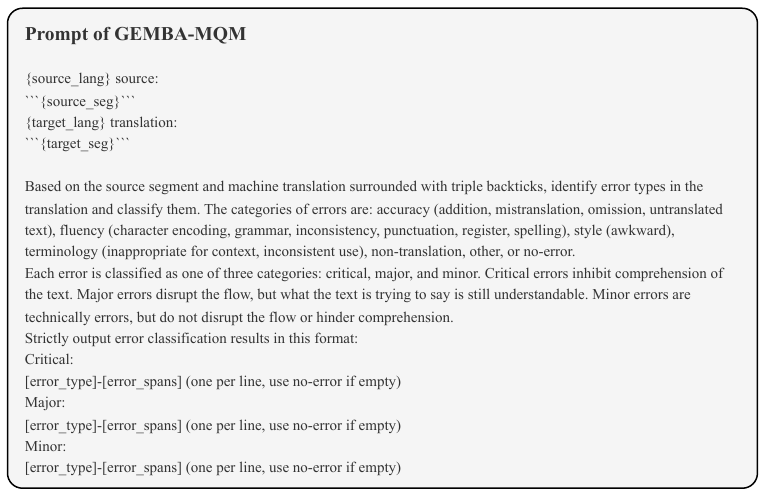}
    \caption{Prompt of GEMBA-MQM, we use GPT-4o as backbone model, and we align it with the original implementation.}
    \label{fig:gembamqm_prompt}
\end{figure*}

\hypertarget{back:gemba_mqm}{\paragraph{GEMBA-MQM \citep{2023-gemba}:}} We employ \texttt{GPT-4o} as the backbone model, aligning our prompt design with the original implementation. 
To maintain consistency in error analysis and computation of final MQM score, we adopt the MQM error span parser from the official implementation\footnote{\url{https://github.com/MicrosoftTranslator/GEMBA/blob/main/gemba/gemba_mqm_utils.py}}.
The detailed prompt used for MQM-style evaluation is illustrated in Figure \ref{fig:gembamqm_prompt}.

\begin{figure*}[t]
    \centering
    \hfill \hyperlink{back:gemba_da}{\textbf{[Back to Text]}}
    \includegraphics[width=0.98\linewidth]{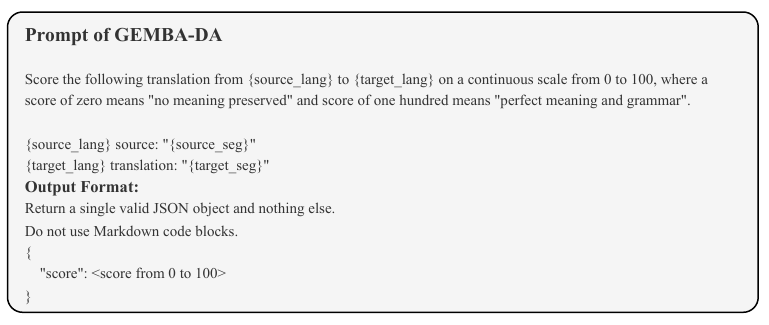}
    \caption{Prompt of GEMBA-DA, we use GPT-4o as backbone model, and we align it with the original implementation, and the only modification is the inclusion of a formatting constraint that explicitly instructs the LLM to output the evaluation score in a structured JSON format, thereby facilitating automated parsing and result extraction accurately.}
    \label{fig:gembada_prompt}
\end{figure*}

\hypertarget{back:gemba_da}{\paragraph{GEMBA-DA \citep{2023-gembada}:}} Utilizing \texttt{GPT-4o} as the backbone model, we align our evaluation with the original GEMBA-DA.  
The only modification involves a structural constraint requiring the model to generate scores in JSON format. This ensures accurately automated parsing and robust extraction of evaluation results. The corresponding prompt is provided in Figure \ref{fig:gembada_prompt}.

\begin{figure*}[t]
    \centering
    \hfill \hyperlink{back:eaprompt}{\textbf{[Back to Text]}}
    \includegraphics[width=0.98\linewidth]{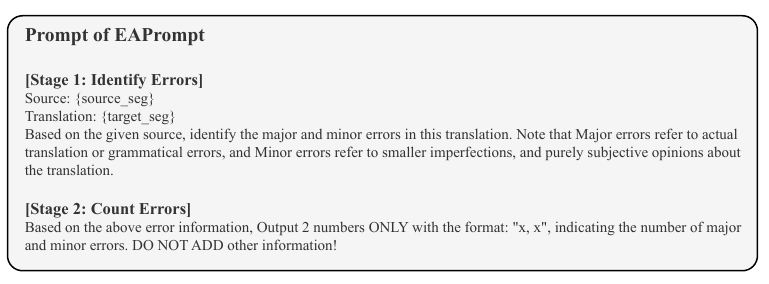}
    \caption{Prompt of EAPrompt, we use GPT-4o as backbone model, and we align it with the original two stages evaluation implementation.}
    \label{fig:eaprompt_prompt}
\end{figure*}

\hypertarget{back:eaprompt}{\paragraph{EAPrompt \citep{2024-eaprompt}:}} We implement the two-stage paradigm (identify errors, count errors) with the backbone model \texttt{GPT-4o}, as proposed in the original implementation of EAPrompt. The two-stage prompt is illustrated in Figure \ref{fig:eaprompt_prompt}.

\paragraph{ThinMQM \citep{2025-thinmqm}:} We evaluate the performance of ThinMQM-32B model\footnote{\url{https://huggingface.co/rzzhan/ThinMQM-32B}}, by strictly adhering to the official prompt template and generation configuration (temperature=0.6, top\_p=0.95)\footnote{\url{https://github.com/NLP2CT/ThinMQM}}.

\paragraph{M-MAD \citep{2025-mad}:} For a fair comparison with other LLM-as-a-Judge paradigms, we use \texttt{GPT-4o} as the backbone model for M-MAD. Except to the selection of backbone model, our evaluation is conducted using the original implementation and official source code\footnote{\url{https://github.com/SU-JIAYUAN/M-MAD}}.

\paragraph{RATE (Ours):} Our proposed agentic translation evaluation framework, detailed implementation of RATE is introduced in Section \ref{sec:rate} and Appendix \ref{sec:appendix_rate_impl}.

\hypertarget{back:domain_eval}{\section{Numerical Meta-Evaluation Results on Each Domain}}
\label{sec:appendix_domain_eval_result}
In this section, we illustrate the numerical meta-evaluation results on each domain in Table \ref{tab:domain_eval_sns}, \ref{tab:domain_eval_culture}, \ref{tab:domain_eval_poetry}, and \ref{tab:domain_eval_literature}, as analyzed in Section \ref{sec:domain_specific_analysis}.

\begin{table*}[htbp]
\hfill \hyperlink{back:domain_eval}{\textbf{[Back to Text]}}
\vspace{5pt}

\centering
\small
\begin{tabular}{l c cccccc cccccc}
\toprule
 & & \multicolumn{6}{c}{\textbf{ZH-EN}} & \multicolumn{6}{c}{\textbf{EN-ZH}} \\
\cmidrule(lr){3-8} \cmidrule(lr){9-14}

\multirow{2}{*}{\textbf{Metric}} & \multirow{2}{*}{\textbf{Meta}} & \multicolumn{3}{c}{System-Level} & \multicolumn{3}{c}{Segment-Level} & \multicolumn{3}{c}{System-Level} & \multicolumn{3}{c}{Segment-Level} \\
\cmidrule(lr){3-5} \cmidrule(lr){6-8} \cmidrule(lr){9-11} \cmidrule(lr){12-14}

 & & Acc. & $r$ & $\rho$ & Acc-t. & $r$ & $\rho$ & Acc. & $r$ & $\rho$ & Acc-t. & $r$ & $\rho$ \\
\midrule

\rowcolor{gray!15}
\multicolumn{14}{l}{\textit{Reference-based Metrics}} \\
BLEU & 69.1 & 91.1 & 94.4 & 93.9 & 57.7 & 32.6 & 36.2 & 91.1 & 92.0 & 93.9 & 57.0 & 44.7 & 44.2 \\
BLEURT & 62.9 & 93.3 & 95.9 & 96.4 & 62.2 & 47.2 & 47.0 & 66.7 & 88.6 & 44.2 & 52.0 & 34.0 & 26.4 \\
COMET & 72.4 & 93.3 & 97.2 & 95.2 & 63.7 & 57.0 & 55.1 & 82.2 & 95.9 & 78.2 & 56.3 & 50.4 & 43.9 \\
XCOMET & 60.1 & 93.3 & 93.7 & 95.2 & 62.4 & 41.5 & 39.4 & 64.4 & 89.3 & 35.8 & 52.0 & 29.1 & 25.4 \\
MetricX-23 & 58.0 & 88.9 & 94.1 & 90.3 & 59.5 & 38.7 & 38.5 & 57.8 & 86.2 & 26.1 & 53.1 & 33.4 & 28.9 \\
MetricX-24 & 59.4 & 86.7 & 90.3 & 84.2 & 60.9 & 51.4 & 50.8 & 53.3 & 88.5 & 21.2 & 55.9 & 35.8 & 33.2 \\
\midrule

\rowcolor{gray!15}
\multicolumn{14}{l}{\textit{Reference-free (QE) Metrics}} \\
COMETKiwi & 40.3 & 48.9 & 71.8 & 1.8 & 48.8 & 28.3 & 27.2 & 53.3 & 80.2 & 10.3 & 50.0 & 35.7 & 26.7 \\
MetricX23-QE & 35.0 & 57.8 & 61.6 & 20.0 & 46.5 & 9.9 & 5.5 & 44.4 & 81.9 & -10.3 & 47.8 & 33.9 & 20.5 \\
MetricX24-QE & 36.5 & 48.9 & 66.2 & 1.8 & 48.5 & 20.3 & 16.1 & 46.7 & 85.1 & -3.0 & 51.5 & 30.4 & 25.6 \\
\midrule

\rowcolor{gray!15}
\multicolumn{14}{l}{\textit{LLM-as-a-Judge}} \\
GEMBA-MQM & 58.9 & 84.4 & 87.0 & 89.1 & 45.7 & 44.2 & 36.1 & 93.3 & 96.9 & 95.2 & 52.3 & 57.9 & 44.0 \\
GEMBA-DA & 75.0 & 91.1 & 84.5 & 95.2 & 55.2 & 56.3 & 58.8 & 91.1 & 95.7 & 95.2 & 57.2 & 62.1 & 57.2 \\ 
EAPrompt & 44.6 & 82.2 & 87.4 & 77.0 & 40.0 & 15.3 & 13.3 & 57.8 & 80.8 & 22.5 & 39.2 & 9.0 & 11.0 \\
ThinMQM & 49.8 & 71.1 & 77.8 & 53.9 & 50.4 & 26.0 & 27.7 & 60.0 & 90.4 & 24.8 & 49.7 & 30.5 & 34.9 \\
M-MAD & 64.7 & 84.4 & 82.3 & 89.4 & 44.8 & 37.6 & 31.7 & 84.4 & 96.8 & 83.0 & 53.2 & 45.1 & 44.0 \\
\textbf{RATE (Ours)} & 80.7 & 93.3 & 99.2 & 96.4 & 63.3 & 66.0 & 66.0 &95.6 & 99.1 & 97.6 & 61.3 & 66.7 & 63.2 \\
\bottomrule
\end{tabular}
\caption{System-level and segment-level correlations on \textbf{SNS domain}. We report \textbf{Accuracy (Acc., Acc-t.)}, \textbf{Pearson($r$)}, and \textbf{Spearman ($\rho$)} correlation coefficients, scaled by a factor of 100. Meta represents the average score of all accuracies and correlation coefficients.}
\label{tab:domain_eval_sns}
\end{table*}

\begin{table*}[htbp]
\centering
\small
\begin{tabular}{l c cccccc cccccc}
\toprule
 & & \multicolumn{6}{c}{\textbf{ZH-EN}} & \multicolumn{6}{c}{\textbf{EN-ZH}} \\
\cmidrule(lr){3-8} \cmidrule(lr){9-14}

\multirow{2}{*}{\textbf{Metric}} & \multirow{2}{*}{\textbf{Meta}} & \multicolumn{3}{c}{System-Level} & \multicolumn{3}{c}{Segment-Level} & \multicolumn{3}{c}{System-Level} & \multicolumn{3}{c}{Segment-Level} \\
\cmidrule(lr){3-5} \cmidrule(lr){6-8} \cmidrule(lr){9-11} \cmidrule(lr){12-14}

 & & Acc. & $r$ & $\rho$ & Acc-t. & $r$ & $\rho$ & Acc. & $r$ & $\rho$ & Acc-t. & $r$ & $\rho$ \\
\midrule

\rowcolor{gray!15}
\multicolumn{14}{l}{\textit{Reference-based Metrics}} \\
BLEU & 65.1 & 86.7 & 86.0 & 87.9 & 50.7 & 40.7 & 32.1 & 86.7 & 88.8 & 89.1 & 57.7 & 38.2 & 36.9 \\
BLEURT & 58.8 & 95.6 & 97.2 & 96.4 & 58.4 & 51.5 & 32.0 & 62.2 & 85.5 & 37.0 & 50.4 & 19.9 & 19.0 \\
COMET & 72.8 & 86.7 & 99.9 & 87.9 & 58.5 & 64.4 & 36.9 & 91.1 & 97.0 & 93.9 & 61.8 & 50.6 & 43.8 \\
XCOMET & 64.3 & 64.4 & 98.6 & 44.2 & 52.5 & 53.9 & 39.0 & 82.2 & 97.5 & 84.2 & 60.8 & 48.7 & 45.4 \\
MetricX-23 & 65.9 & 68.9 & 98.3 & 58.8 & 55.1 & 57.4 & 36.8 & 82.2 & 97.1 & 75.8 & 59.5 & 51.8 & 49.1 \\
MetricX-24 & 62.8 & 66.7 & 98.6 & 52.7 & 55.4 & 57.7 & 35.4 & 71.1 & 95.0 & 53.9 & 60.3 & 56.5 & 49.5 \\
\midrule

\rowcolor{gray!15}
\multicolumn{14}{l}{\textit{Reference-free (QE) Metrics}} \\
COMETKiwi & 45.8 & 48.9 & 96.6 & -4.2 & 51.9 & 49.9 & 29.2 & 51.1 & 89.5 & 11.5 & 53.8 & 41.2 & 29.6 \\
MetricX23-QE & 45.3 & 53.3 & 96.2 & 6.7 & 52.1 & 54.6 & 32.5 & 46.7 & 85.5 & -4.2 & 51.8 & 36.3 & 31.8 \\
MetricX24-QE & 52.4 & 53.3 & 97.4 & 11.5 & 52.7 & 57.8 & 31.1 & 62.2 & 90.0 & 30.9 & 54.8 & 47.0 & 40.0 \\
\midrule

\rowcolor{gray!15}
\multicolumn{14}{l}{\textit{LLM-as-a-Judge}} \\
GEMBA-MQM & 70.1 & 77.8 & 99.0 & 75.8 & 54.0 & 74.1 & 41.3 & 82.2 & 97.6 & 84.2 & 52.3 & 58.2 & 45.1 \\
GEMBA-DA & 75.0 & 82.2 & 99.5 & 80.6 & 60.0 & 83.3 & 50.3 & 86.7 & 97.2 & 89.1 & 50.6 & 61.6 & 58.5 \\ 
EAPrompt & 51.0 & 68.9 & 95.0 & 58.8 & 40.7 & 20.8 & 17.4 & 73.3 & 95.1 & 64.4 & 41.2 & 21.7 & 14.4 \\
ThinMQM & 68.9 & 80.0 & 99.1 & 75.8 & 51.9 & 66.0 & 36.2 & 86.7 & 96.6 & 86.7 & 51.4 & 50.4 & 46.4 \\
M-MAD & 67.8 & 84.4 & 99.5 & 80.6 & 48.9 & 60.8 & 36.9 & 80.0 & 96.6 & 78.2 & 51.4 & 50.9 & 45.6 \\
\textbf{RATE (Ours)} & 78.2 & 86.7 & 98.9 & 85.5 & 60.6 & 80.2 & 54.3 & 95.6 &99.4 & 97.6 & 57.0 & 63.8 & 58.1 \\
\bottomrule
\end{tabular}
\caption{System-level and segment-level correlations on \textbf{Cross-Culture domain}. We report \textbf{Accuracy (Acc., Acc-t.)}, \textbf{Pearson($r$)}, and \textbf{Spearman ($\rho$)} correlation coefficients, scaled by a factor of 100. Meta represents the average score of all accuracies and correlation coefficients.}
\label{tab:domain_eval_culture}
\end{table*}

\begin{table*}[t]
\hfill \hyperlink{back:domain_eval}{\textbf{[Back to Text]}}
\vspace{5pt}

\centering
\small
\begin{tabular}{l c cccccc cccccc}
\toprule
 & & \multicolumn{6}{c}{\textbf{ZH-EN}} & \multicolumn{6}{c}{\textbf{EN-ZH}} \\
\cmidrule(lr){3-8} \cmidrule(lr){9-14}

\multirow{2}{*}{\textbf{Metric}} & \multirow{2}{*}{\textbf{Meta}} & \multicolumn{3}{c}{System-Level} & \multicolumn{3}{c}{Segment-Level} & \multicolumn{3}{c}{System-Level} & \multicolumn{3}{c}{Segment-Level} \\
\cmidrule(lr){3-5} \cmidrule(lr){6-8} \cmidrule(lr){9-11} \cmidrule(lr){12-14}

 & & Acc. & $r$ & $\rho$ & Acc-t. & $r$ & $\rho$ & Acc. & $r$ & $\rho$ & Acc-t. & $r$ & $\rho$ \\
\midrule

\rowcolor{gray!15}
\multicolumn{14}{l}{\textit{Reference-based Metrics}} \\
BLEU & 54.8 & 75.6 & 26.8 & 51.5 & 47.8 & 5.6 & 13.5 & 93.3 & 95.2 & 95.2 & 64.8 & 41.6 & 46.0 \\
BLEURT & 47.4 & 66.7 & 86.3 & 43.0 & 54.8 & 35.5 & 30.6 & 53.3 & 66.9 & 17.6 & 53.6 & 32.0 & 27.9 \\
COMET & 61.8 & 71.1 & 89.2 & 60.0 & 53.5 & 40.6 & 29.2 & 77.8 & 85.2 & 62.4 & 63.6 & 58.0 & 50.3 \\
XCOMET & 54.8 & 77.8 & 92.3 & 63.6 & 51.3 & 35.3 & 25.5 & 64.4 & 77.0 & 33.3 & 57.0 & 44.0 & 36.0 \\
MetricX-23 & 44.9 & 53.3 & 76.7 & 18.8 & 47.7 & 32.9 & 21.4 & 60.0 & 71.2 & 22.4 & 54.1 & 47.5 & 32.5 \\
MetricX-24 & 49.3 & 62.2 & 82.4 & 33.3 & 51.7 & 44.0 & 31.3 & 60.0 & 70.1 & 16.4 & 54.6 & 51.2 & 34.7 \\
\midrule

\rowcolor{gray!15}
\multicolumn{14}{l}{\textit{Reference-free (QE) Metrics}} \\
COMETKiwi & 52.0 & 71.1 & 88.2 & 61.2 & 52.8 & 38.1 & 28.9 & 57.8 & 70.6 & 16.4 & 54.8 & 47.4 & 36.5 \\
MetricX23-QE & 36.9 & 53.3 & 49.6 & 18.8 & 44.6 & 18.8 & 12.2 & 53.3 & 68.2 & 6.7 & 48.5 & 43.3 & 25.5 \\
MetricX24-QE & 40.5 & 53.3 & 70.2 & 18.8 & 48.7 & 33.4 & 21.5 & 51.1 & 63.1 & 4.2 & 50.4 & 44.3 & 26.4 \\
\midrule

\rowcolor{gray!15}
\multicolumn{14}{l}{\textit{LLM-as-a-Judge}} \\
GEMBA-MQM & 71.9 & 86.7 & 95.9 & 87.9 & 44.8 & 64.1 & 44.0 & 91.1 & 88.3 & 92.7 & 49.6 & 64.9 & 51.8 \\
GEMBA-DA & 77.9 & 84.4 & 94.0 & 86.7 & 59.3 & 73.3 & 60.5 & 93.3 & 89.6 & 95.2 & 59.3 & 72.0 & 67.4  \\ 
EAPrompt & 55.3 & 88.9 & 92.0 & 90.3 & 42.0 & 28.4 & 21.1 & 73.3 & 84.8 & 55.2 & 43.1 & 23.4 & 20.3 \\
ThinMQM & 71.8 & 88.9 & 95.5 & 89.1 & 54.5 & 57.9 & 44.7 & 84.4 & 83.6 & 83.0 & 51.8 & 56.5 & 47.3 \\
M-MAD & 67.6 & 84.4 & 93.3 & 80.6 & 46.8 & 52.7 & 36.0 & 88.9 & 88.8 & 90.3 & 50.3 & 50.8 & 48.2 \\
\textbf{RATE (Ours)} & 76.7 & 88.9 & 97.4 & 91.2 & 63.2 & 70.2 & 61.5 & 86.7 & 93.8 & 86.7 & 58.6 & 61.8 & 56.1 \\
\bottomrule
\end{tabular}
\caption{System-level and segment-level correlations on \textbf{Poetry domain}. We report \textbf{Accuracy (Acc., Acc-t.)}, \textbf{Pearson($r$)}, and \textbf{Spearman ($\rho$)} correlation coefficients, scaled by a factor of 100. Meta represents the average score of all accuracies and correlation coefficients.}
\label{tab:domain_eval_poetry}
\end{table*}

\begin{table*}[t]
\centering
\small
\begin{tabular}{l c cccccc cccccc}
\toprule
 & & \multicolumn{6}{c}{\textbf{ZH-EN}} & \multicolumn{6}{c}{\textbf{EN-ZH}} \\
\cmidrule(lr){3-8} \cmidrule(lr){9-14}

\multirow{2}{*}{\textbf{Metric}} & \multirow{2}{*}{\textbf{Meta}} & \multicolumn{3}{c}{System-Level} & \multicolumn{3}{c}{Segment-Level} & \multicolumn{3}{c}{System-Level} & \multicolumn{3}{c}{Segment-Level} \\
\cmidrule(lr){3-5} \cmidrule(lr){6-8} \cmidrule(lr){9-11} \cmidrule(lr){12-14}

 & & Acc. & $r$ & $\rho$ & Acc-t. & $r$ & $\rho$ & Acc. & $r$ & $\rho$ & Acc-t. & $r$ & $\rho$ \\
\midrule

\rowcolor{gray!15}
\multicolumn{14}{l}{\textit{Reference-based Metrics}} \\
BLEU & 67.0 & 80.0 & 87.2 & 79.4 & 51.6 & 39.4 & 32.2 & 88.9 & 94.7 & 89.1 & 60.4 & 44.4 & 44.1 \\
BLEURT & 57.8 & 82.2 & 97.6 & 80.6 & 58.0 & 51.5 & 40.1 & 62.2 & 82.1 & 30.9 & 51.3 & 33.8 & 23.2 \\
COMET & 66.6 & 86.7 & 99.1 & 83.0 & 58.8 & 60.7 & 45.0 & 71.1 & 85.1 & 46.7 & 55.9 & 64.4 & 42.9 \\
XCOMET & 53.9 & 68.9 & 96.5 & 60.0 & 54.1 & 49.5 & 34.6 & 66.7 & 75.4 & 34.5 & 53.4 & 29.7 & 22.8 \\
MetricX-23 & 56.3 & 75.6 & 96.3 & 69.7 & 55.5 & 64.3 & 37.4 & 55.6 & 77.0 & 16.4 & 50.2 & 51.4 & 25.9 \\
MetricX-24 & 56.1 & 71.1 & 96.7 & 66.1 & 53.6 & 64.9 & 39.2 & 55.6 & 78.0 & 16.4 & 49.2 & 56.6 & 26.3 \\
\midrule

\rowcolor{gray!15}
\multicolumn{14}{l}{\textit{Reference-free (QE) Metrics}} \\
COMETKiwi & 43.2 & 53.3 & 93.6 & 12.7 & 48.4 & 42.5 & 24.7 & 44.4 & 75.6 & -7.9 & 48.8 & 49.8 & 31.9 \\
MetricX23-QE & 46.3 & 68.9 & 92.0 & 47.9 & 49.2 & 52.9 & 27.5 & 37.8 & 77.4 & -21.2 & 45.3 & 57.0 & 21.1 \\
MetricX24-QE & 48.0 & 62.2 & 94.4 & 41.8 & 49.5 & 54.2 & 29.7 & 48.9 & 75.9 & -0.6 & 46.7 & 53.2 & 19.7 \\
\midrule

\rowcolor{gray!15}
\multicolumn{14}{l}{\textit{LLM-as-a-Judge}} \\
GEMBA-MQM & 68.2 & 68.9 & 97.2 & 53.9 & 50.2 & 64.4 & 32.3 & 93.3 & 98.6 & 95.2 & 53.1 & 61.8 & 48.5 \\
GEMBA-DA & 77.6 & 80.0 & 97.3 & 77.0 & 56.4 & 77.4 & 50.3 & 91.1 & 92.6 & 93.9 & 65.7 & 79.4 & 69.4 \\ 
EAPrompt & 45.2 & 66.7 & 76.6 & 40.6 & 38.0 & 9.3 & 3.89 & 80.0 & 92.8 & 69.7 & 42.6 & 14.1 & 7.7 \\
ThinMQM & 61.0 & 60.0 & 96.1 & 34.5 & 50.2 & 44.8 & 33.7 & 82.2 & 90.7 & 83.0 & 55.3 & 53.4 & 47.5 \\
M-MAD & 66.6 & 75.6 & 96.6 & 67.3 & 47.5 & 44.8 & 26.8 & 91.1 & 94.3 & 91.5 & 55.4 & 57.8 & 50. 1  \\
\textbf{RATE (Ours)} & 78.9 & 88.9 & 99.1 & 87.9 & 60.8 & 77.4 & 60.4 & 88.9 & 95.2 & 92.7 & 62.1 & 69.8 & 63.6\\
\bottomrule
\end{tabular}
\caption{System-level and segment-level correlations on \textbf{Literature domain}. We report \textbf{Accuracy (Acc., Acc-t.)}, \textbf{Pearson($r$)}, and \textbf{Spearman ($\rho$)} correlation coefficients, scaled by a factor of 100. Meta represents the average score of all accuracies and correlation coefficients.}
\label{tab:domain_eval_literature}
\end{table*}

\end{document}